%% file: main.tex
\documentclass[10pt,twocolumn,letterpaper]{article}

\usepackage[pagenumbers]{cvpr} 

\input{preamble}
\definecolor{cvprblue}{rgb}{0.21,0.49,0.74}
\usepackage[pagebackref,breaklinks,colorlinks,allcolors=cvprblue]{hyperref}

\usepackage{multirow}

\usepackage{wrapfig}

\usepackage[accsupp]{axessibility} 

\usepackage{xspace}
\newcommand{\tiger}{\texttt{TIGeR}\xspace}

\usepackage{tikz}
\def\checkmark{\tikz\fill[scale=0.4](0,.35) -- (.25,0) -- (1,.7) -- (.25,.15) -- cycle;} 

\DeclareMathOperator*{\argmax}{arg\,max}

\def\paperID{37541} 
\def\confName{CVPR}
\def\confYear{2026}

\title{\tiger: A Unified Framework for Time, Images and Geo-location Retrieval}

\author{David G. Shatwell \quad Sirnam Swetha  \quad Mubarak Shah \\
Institute of Artificial Intelligence, University of Central Florida \\
{\tt\small david.shatwell@ucf.edu \quad swetha.sirnam@ucf.edu \quad shah@crcv.ucf.edu}
}

\begin{document}
\maketitle
\input{sec/0_abstract}

\input{sec/1_intro}
\input{sec/2_background}
\input{sec/4_method}

\input{sec/3_dataset}
\input{sec/5_exp}

\input{sec/6_conclusion}

\input{sec/7_acknowledgement}
{
    \small
    \bibliographystyle{ieeenat_fullname}
    \bibliography{main}
}

\input{sec/X_suppl}

\end{document}

%% file: sec/0_abstract.tex
\begin{abstract}

Many real-world applications in digital forensics, urban monitoring, and environmental analysis require jointly reasoning about visual appearance, location, and time. Beyond standard geo-localization and time-of-capture prediction, these applications increasingly demand more complex capabilities, such as retrieving an image captured at the same location as a query image but at a specified target time. We formalize this problem as \emph{Geo-Time Aware Image Retrieval} and propose \tiger, a unified framework for \textbf{T}ime, \textbf{I}mages and \textbf{Ge}o-location \textbf{R}etrieval. \tiger supports flexible input configurations (single-modality and multi-modality queries) and uses the same representation to perform (i) geo-localization, (ii) time-of-capture prediction, and (iii) geo-time–aware retrieval. By preserving the underlying location identity despite large appearance changes, \tiger enables retrieval based on where and when a scene was captured, rather than purely on visual similarity. To support this task, we design a multistage data curation pipeline and propose a new diverse dataset of 4.5M paired image–location–time triplets for training and 86k high-quality triplets for evaluation. Extensive experiments show that \tiger consistently outperforms strong baselines and state-of-the-art methods by up to 16\% on time-of-year, 8\% time-of-day prediction, and 14\% in geo-time aware retrieval recall, highlighting the benefits of unified geo-temporal modeling.

\end{abstract}

%% file: sec/1_intro.tex
\vspace{-1em}

\section{Introduction}

Understanding how a location visually appears across different points in time is essential for many real-world applications, including digital forensics, environmental and climate change analysis, cultural heritage preservation, and augmented reality. In these settings, the goal is not simply to retrieve images that look similar to a query, but to retrieve images that correspond to the \emph{same physical location} at a \emph{specified target time}. For example, a query image of a city square in winter should retrieve an image of the same square in summer, despite large differences in foliage, lighting, pedestrian patterns, or ongoing construction. This capability requires models that represent location identity robustly while explicitly accounting for temporal variation in how that location appears.

\begin{figure}[t]
\begin{center}
\includegraphics[width=0.9\linewidth]{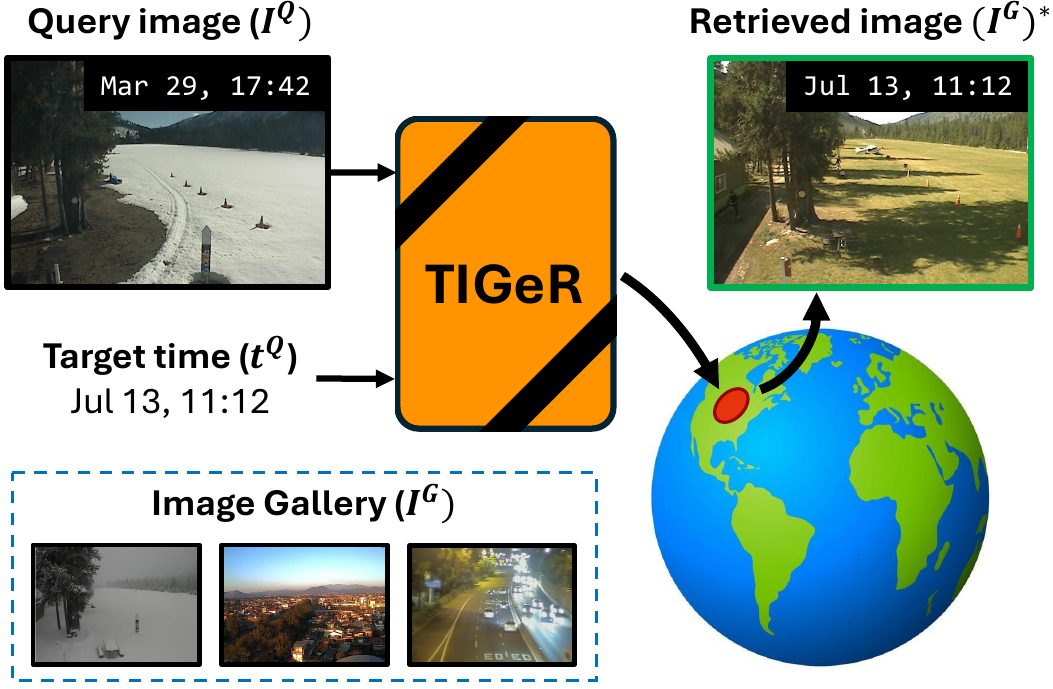}
\end{center}
\vspace{-1em}
\caption{
\tiger unifies time, image, and location understanding, enabling geo-localization, time prediction, and geo-time aware image retrieval: given a query image and a target time, it retrieves an image captured at the same location at the specified time.
}
\label{fig:geotime-fig}
\end{figure}

We formalize this setting as \emph{Geo-Time Aware Image Retrieval}. Given a query image and a target time, the task is to retrieve an image captured at the same location at the specified time, rather than one chosen only based on visual similarity (Figure~\ref{fig:geotime-fig}). The central challenge is to learn representations that factor out appearance changes driven by time while preserving the underlying geo-location semantics of a place. 
However, existing approaches do not address this challenge. Standard image retrieval methods primarily rank images based on appearance similarity \cite{zhu2023r2former, keetha2023anyloc, Berton_2025_CVPR}, but are trained to be invariant to the time-of-capture. Composed retrieval methods can modify visual attributes (e.g., “with snow,” “at night”) \cite{saito2023pic2word, baldrati2023zero}, but these transformations operate purely in the image domain and do not guarantee that the retrieved image comes from the same geographic location. Geo-localization models, on the other hand, aim to estimate where an image was taken, and prior work \cite{zhai2019learning, Shatwell_2025_ICCV} extends this to jointly infer where and when an image was captured. Yet these models typically encode each modality (image, location, time) independently and align only the final embeddings through contrastive learning. Without an explicit cross-modal fusion mechanism, they struggle to capture rich geo-temporal relationships and to model how visual appearance at a fixed location evolves over time.

To address this challenge, we propose a novel approach using a multi-modal transformer that leverages self-attention to jointly process visual, geo-location, and temporal information. One or more modality-specific embeddings are passed through the multi-modal transformer, aggregating information across modalities and producing a unified representation that is then projected into a joint geo-temporal embedding space. By allowing modalities to attend to each other, the model learns geo-temporal correlations \emph{directly}, rather than aligning separately learned embeddings post hoc, thereby enabling robust retrieval even when metadata is incomplete or noisy.

The learned representation serves as a unified geo-temporal embedding that supports: (i) geo-localization, (ii) time-of-capture prediction, and (iii) geo-time aware image retrieval. Given an image and a target time, \tiger performs retrieval based on \emph{where} and \emph{when} the scene is, rather than purely on how it looks, making it more robust to substantial seasonal, structural, and lighting changes.

However, developing such models also requires large, high-quality image datasets paired with geo-locations and timestamps. Existing datasets are often geographically biased toward the Northern Hemisphere \cite{mihail2016sky} or contain substantial degraded or low-quality imagery \cite{jacobs07amos, jacobs09webcamgis}, making them insufficient for rigorous evaluation at global scale. To address this gap, we introduce a new benchmark for geo-time aware image retrieval. Our benchmark is constructed from a curated subset of the AMOS webcam corpus~\cite{jacobs09webcamgis, jacobs07amos}, which offers global geographic coverage and continuous image capture across seasons, weather conditions, and times of day. We develop a multi-stage curation pipeline to (i) remove noisy, low-quality, or corrupted frames, (ii) normalize and balance geographic coverage, and (iii) construct train/test splits that preserve both spatial and temporal diversity. The resulting dataset provides a challenging yet reliable platform for evaluating models that must reason jointly over images, locations, and timestamps.


\noindent In summary, our key contributions are as follows:
\begin{itemize}
    \item We introduce geo-time aware image retrieval, a new task that requires retrieving images consistent with both the geographic location and target time-of-capture of a query.
    \item We propose \tiger, a multimodal transformer-based framework that learns a unified geo-temporal embedding space through cross-modal attention over image, location, and time inputs.
    \item We curate both a large-scale training dataset and a benchmark dataset with global geographic coverage, broad temporal diversity, and extensive quality filtering, providing the first large-scale testbed tailored to this task.
    \item We show that \tiger outperforms prior methods, achieving average improvements of 16\% on time-of-year prediction, 8\% on time-of-day prediction, and up to 14\% over state-of-the-art baselines for time-of-capture prediction and geo-time aware retrieval.
\end{itemize}

%% file: sec/2_background.tex
\section{Related Work}
\label{sec:rel}

\textbf{Time-aware understanding.}
Modeling how visual appearance evolves over time has recently attracted growing attention in computer vision.
Early works explored predicting the \textit{time-of-capture} from single images using either physical or data-driven cues.
Physically inspired methods~\citep{tsai2016photo, li2017you} estimate time or verify timestamp consistency by analyzing shadows or the Sun’s position, but they require strong priors such as known camera orientation, sky visibility, or external metadata.
Data-centric approaches~\citep{zhai2019learning, salem2022timestamp, padilha2022content} instead train neural networks to predict temporal attributes such as time-of-day or month from labeled images, often in combination with geo-tags.
However, these models typically treat time as a classification target and are limited to coarse, discretized predictions.
GT-Loc~\citep{Shatwell_2025_ICCV} extends this direction by learning a shared embedding space for image, location, and time, but processes each modality independently and aligns them only via a contrastive loss at the embedding level, restricting its ability to capture rich cross-modal geo-temporal relationships.
In contrast, \tiger explicitly learns these relationships by enabling cross-modal attention between modalities, producing a unified representation that models how visual appearance varies jointly across geolocation and time.

\textbf{Global geo-localization.}
Geo-localization aims to infer the geographic coordinates of an image or retrieve images taken at similar locations from a large gallery.
Traditional methods either discretize the globe into geo-cells and frame localization as classification~\citep{weyand2016planet, vo2017revisiting, muller2018geolocation, pramanick2022world, kulkarni2024cityguessr}, or adopt a retrieval-based paradigm that matches query embeddings to stored geospatial features~\citep{regmi2019bridging, shi2020looking, zhu2021vigor, zhu2022transgeo, vivanco2024geoclip, Shatwell_2025_ICCV}.
Recent approaches such as GeoCLIP~\citep{vivanco2024geoclip} leverage CLIP-style contrastive training to align image and GPS embeddings in a shared feature space, improving large-scale retrieval.
PIGEON~\citep{Haas_2024_CVPR} combines classification and retrieval for fine-grained localization, while Img2Loc~\citep{zhou2024img2loc} and GAEA~\citep{campos2025gaea} use multimodal large language models to reason jointly over image content and metadata.
However, these approaches still operate with largely independent encoders and do not explicitly model cross-modal relationships governing geo-temporal variation.
\tiger bridges this gap by using attention-based fusion to learn a shared representation space in which both \textit{where} and \textit{when} matter: the model can retrieve images captured at the \textit{same} location but at a \textit{different} time, enabling genuine geo-temporal understanding.

\textbf{Geo-temporal and multi-modal representation learning.}
Beyond standard geo-localization, several works seek to learn representations that combine multiple modalities such as images, satellite, and GPS.
SatCLIP~\citep{klemmer2023satclip} and related approaches~\citep{zavras2024mind, pmlr-v202-mai23a, Aodha_2019_ICCV} extend CLIP-like dual encoders approach to align satellite, ground-level views, or GPS for cross-domain geospatial tasks.
While these dual-encoder methods highlight the effectiveness of contrastive learning for spatial alignment, they still encode each modality independently and rely on fixed modality pairs.
\tiger departs from this paradigm by employing a shared multimodal transformer and joint contrastive learning that allows image, GPS, and time to \emph{attend to each other}, inspired by~\cite{shvetsova2022everything, swetha2023preserving}.
This design enables the model to learn fine-grained geo-temporal correlations (e.g., how a specific location changes across seasons or lighting conditions) and to perform retrieval across arbitrary modality combinations, rather than being constrained to pre-defined pairs.

%% file: sec/4_method.tex
\section{Method}

\begin{figure*}[t]
\begin{center}
\includegraphics[
    width=0.925\linewidth,
    trim=0.75cm 0.5cm 0.75cm 0.5cm,
    clip
]{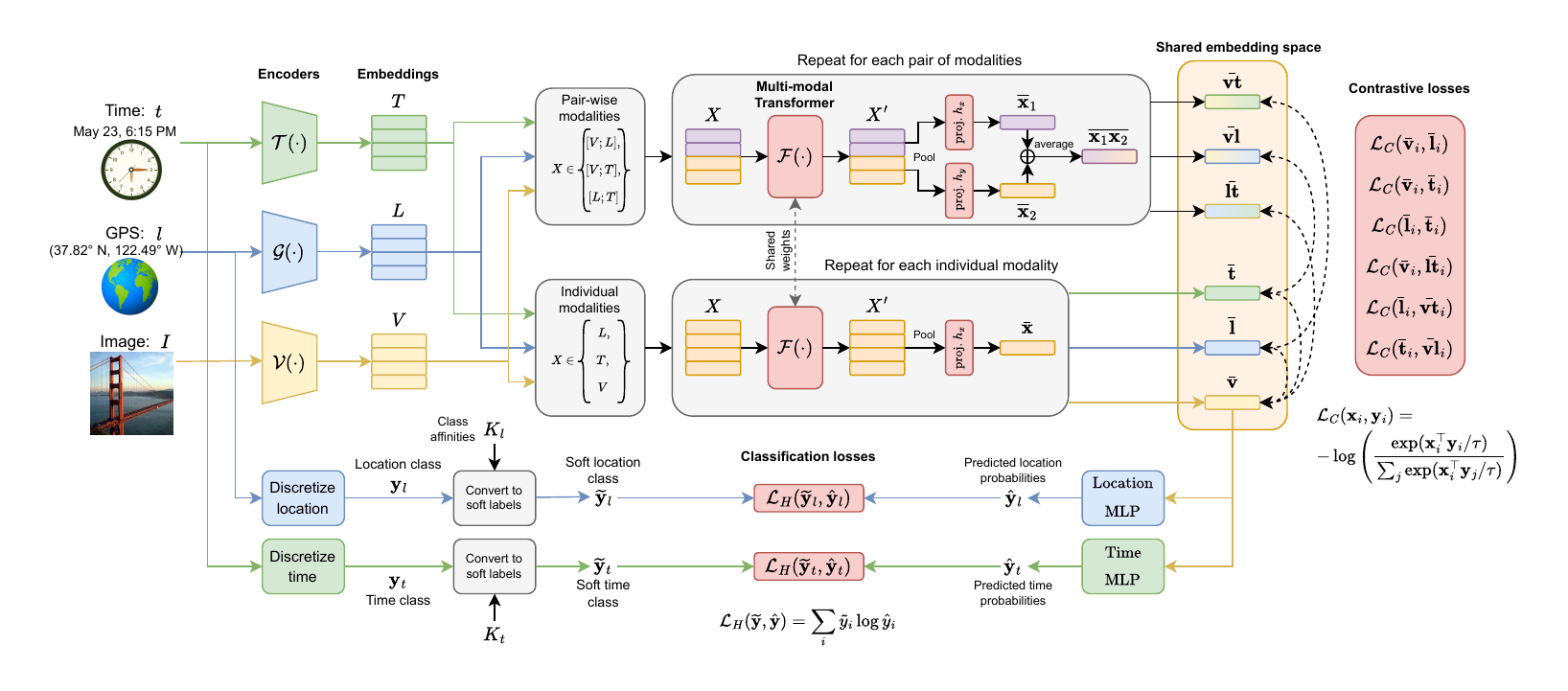}
\end{center}
\vspace{-2em}
\caption{\textbf{Architecture of \tiger.} Given an image $I$, GPS coordinates $l$, and timestamp $t$, modality-specific encoders produce visual ($V$), location ($L$), and temporal ($T$) embeddings. 
    Individual and pair-wise modality tokens are passed through a \textit{shared multi-modal transformer} $\mathcal{F}(\cdot)$, which employs self-attention to learn geo-temporal interactions. 
    Pooled embeddings are projected into a \textit{shared embedding space}, where a \textbf{contrastive loss} aligns unimodal and fused representations across all modality pairings (e.g., $V$--$L$, $V$--$T$, $L$--$T$). In parallel, the \textbf{classification losses with metric targets} on discretized location and time classes encourage the model to learn hierarchical cross-modal representations that vary smoothly across the embedding space.
\
}
\label{fig:architecture}
\end{figure*}

Our goal is to learn a unified representation that jointly embeds the location, visual, and time modalities, namely, the image $I$, GPS coordinates $l$, and time $t$ into a shared geo-temporal feature space. The representation is designed to preserve location identity while capturing temporal variation, enabling flexible cross-modal retrieval. Our proposed model is capable of  various tasks, such as, (i) retrieve an image of the same place at a different time, (ii) estimate the geo-location from image alone, and (iii) predict a plausible time-of-capture from single image or (iv) combined image and location cues. 

\subsection{Problem Statement}

Given a dataset $S_{\mathrm{train}}=\{(I_i, l_i, t_i)\}_{i=1}^N$, where $I_i$ is an image, $l_i$ denotes its GPS coordinates, and $t_i$ its capture time, our goal is to learn a model $f:\mathcal{M}\!\rightarrow\!\mathbb{R}^d$ that maps an input $\mathcal{M}$ from the modality set $\{I, l, t, (I,l), (I,t), (l,t)\}$ into a shared feature space $\mathbb{R}^d$. The model is trained so that features from inputs describing similar geo-temporal conditions (\textit{i.e.}, captured at nearby locations and times) have high cosine similarity, while unrelated inputs are pushed apart.

\subsection{Architecture}

Our model comprises three stages (see  Figure \ref{fig:architecture}). The first stage includes an image encoder $\mathcal{V}(I)=V\!\in\!\mathbb{R}^{N_V\times d}$, a location encoder $\mathcal{G}(l)=L\!\in\!\mathbb{R}^{N_L\times d}$, and a time encoder $\mathcal{T}(t)=T\!\in\!\mathbb{R}^{N_T\times d}$, where $N_V$, $N_L$, and $N_T$ denote the number of tokens per modality and $d$ the embedding dimension. The image encoder is a frozen CLIP ViT \cite{clip} that outputs one \texttt{[CLS]} token and $N_V\!-\!1$ patch embeddings. The location and time encoders use random Fourier features (RFF) to project 2D inputs into $N_L$ and $N_T$ high-dimensional vectors, modulated by frequencies $\sigma_i\!\in\!\{2^{2i}\}_{i=0}^{N-1}$. Each modality embedding is then linearly projected and layer-normalized.

The second stage employs a multi-modal transformer $\mathcal{F}$ to refine the modality-specific embeddings, learning fine-grained relationships between them via self-attention. Since there are six possible input combinations, $\{V,L,T,[V;L], [V;T], [L;T]\}$, we perform six forward passes during training. When two modalities are provided, we concatenate their embeddings along the token dimension before fusion, i.e., $X\!\in\!\{V,L,T,[V;L],[V;T],[L;T]\}$ and obtain $X'=\mathcal{F}(X)$.

For each refined embedding $X'$, we perform pooling to aggregate token representations and then project them into the shared embedding space via learned projections $h_x(\cdot)$:
\[
\mathbf{x}=\tfrac{1}{N_x}\sum_{i=1}^{N_x}X'_{i,:},\quad 
\mathbf{x}'=h_x(\mathbf{x}),\quad 
f(x)=\bar{\mathbf{x}}=\tfrac{\mathbf{x}'}{\|\mathbf{x}'\|},
\]
Similarly for bimodal inputs, we first pool and project each constituent modality to obtain embeddings $\bar{\mathbf{x}}_1$ and $\bar{\mathbf{x}}_2$, 
following the same procedure used for unimodal inputs. The resulting normalized embeddings are then fused by averaging: 
\[
f(x_1, x_2) = 
\frac{\bar{\mathbf{x}}_1 + \bar{\mathbf{x}}_2}{
\|\bar{\mathbf{x}}_1 + \bar{\mathbf{x}}_2\|}.
\]
This yields six embeddings $\{\bar{\mathbf{v}}, \bar{\mathbf{l}}, \bar{\mathbf{t}}, \bar{\mathbf{vl}}, \bar{\mathbf{vt}}, \bar{\mathbf{lt}}\}$, all residing in a shared geo-temporal feature space.

\subsection{Losses}

\tiger is optimized using a combination of contrastive and classification losses with soft metric targets. The contrastive objectives encourage embeddings from different input modalities to align within the shared feature space, forming the foundation for all geo-temporal retrieval tasks. In parallel, a classification loss is applied only to the final image embeddings $\bar{\mathbf{v}}$, enforcing a smooth, structured organization of the embedding space guided by the soft metric targets.

\noindent\textbf{Contrastive losses.}
We employ five contrastive objectives to align all meaningful modality pairs, excluding the individual location and time embeddings $(\bar{\mathbf{l}},\bar{\mathbf{t}})$, which remain uncorrelated. Specifically, we optimize the pairs $(\bar{\mathbf{v}},\bar{\mathbf{l}})$, $(\bar{\mathbf{v}},\bar{\mathbf{t}})$, $(\bar{\mathbf{v}},\bar{\mathbf{lt}})$, $(\bar{\mathbf{l}},\bar{\mathbf{vt}})$, and $(\bar{\mathbf{t}},\bar{\mathbf{vl}})$. Each pair $(\mathbf{x}_i,\mathbf{y}_i)$ is trained using the InfoNCE loss~\cite{oord2018representation}:

\begin{equation}
\mathcal{L}_C^{(i)}(\mathbf{x}_i,\mathbf{y}_i)
= -\log \frac{\exp(\mathbf{x}_i^\top \mathbf{y}_i / \tau)}{\sum_j \exp(\mathbf{x}_i^\top \mathbf{y}_j / \tau)},
\end{equation}
where $\tau$ is a temperature parameter controlling the sharpness of the distribution. The overall contrastive objective is the sum of all pairwise components:
\begin{equation}
\begin{aligned}
\mathcal{L}_C
= &\; \mathcal{L}_C(\bar{\mathbf{v}},\bar{\mathbf{l}})
+ \mathcal{L}_C(\bar{\mathbf{v}},\bar{\mathbf{t}})
+ \mathcal{L}_C(\bar{\mathbf{v}},\bar{\mathbf{lt}}) \\
&+ \mathcal{L}_C(\bar{\mathbf{l}},\bar{\mathbf{vt}})
+ \mathcal{L}_C(\bar{\mathbf{t}},\bar{\mathbf{vl}}).
\end{aligned}
\label{eq:contrastive-losses}
\end{equation}

\noindent\textbf{Classification losses.}
To encourage the model to learn geo-temporal representations that vary smoothly across the embedding space, we employ two classification losses with soft targets on the image embedding $\bar{\mathbf{v}}$.

\textit{Geo-location classes.}  
We define spatial classes using the Hierarchical Equal Area isoLatitude Pixelisation (HEALPix) \cite{gorski2005healpix} scheme, which divides the Earth's surface into equal-area regions whose centers lie on discretized latitude circles. The base resolution yields 12 cells, and each subdivision increases the count fourfold. We use the fourth hierarchy level ($12\times4^3=768$ regions). For each image–GPS pair $(I,l)$, we compute the HEALPix label as $(\mathbf{y}_l,l_C)=\mathrm{HEALPix}(l)$, where $\mathbf{y}_l$ is the one-hot class vector and $l_C$ the coordinates of the corresponding cell center.

\textit{Time classes.}  
Temporal classes are defined by binning the time-of-day (ToD) and time-of-year (ToY) into 1-hour and 1-month intervals, producing $24\times12=288$ bins. Following GT-Loc \cite{Shatwell_2025_ICCV}, each timestamp is represented as a point on the flat torus $t=(\theta,\phi)\in\mathbb{T}^2$, partitioned into 288 equal-area regions. For each $(I,t)$ pair, we assign $(\mathbf{y}_t,t_C)=\mathrm{TimeClass}(t)$, where $\mathbf{y}_t$ is the one-hot class vector and $t_C$ the center of the corresponding region.

\textit{Soft targets.}  
To obtain soft supervision, we measure the similarity between class centers via a metric $\kappa$. For geo-location classes, $\kappa$ is the Haversine (great-circle) distance on the sphere. For time classes, we use the geodesic distance on the torus:
\begin{equation}
    \kappa(t_i,t_j)=\sqrt{\sum_{\alpha\in\{\theta,\phi\}}
    \min\!\big(1-|\Delta\alpha_{i,j}|,|\Delta\alpha_{i,j}|\big)^2 }.
\end{equation}
Given a temperature $\gamma$, we construct a normalized affinity matrix
\begin{equation}
    K_{i,j}=\frac{\exp[-\kappa(C_i,C_j)/\gamma]}
    {\sum_j\exp[-\kappa(C_i,C_j)/\gamma]},
\end{equation}
where $C_i$ and $C_j$ are the centers of two classes of the same modality.  
The soft target is then $\widetilde{\mathbf{y}}=K\mathbf{y}$, which spreads probability mass over neighboring classes. The final classification loss is the cross-entropy between the predicted probabilities
$\hat{\mathbf{y}}=\sigma(\mathrm{MLP}(\bar{\mathbf{v}}))$
and the soft target:
\begin{equation}
    \mathcal{L}_H=-\sum_i \widetilde{y}_i\log\hat{y}_i.
\end{equation}

%% file: sec/3_dataset.tex
\section{Benchmark Dataset Curation}
\label{sec:data_curation}

\begin{figure*}[t!]
\begin{center}
\includegraphics[width=0.95\textwidth]{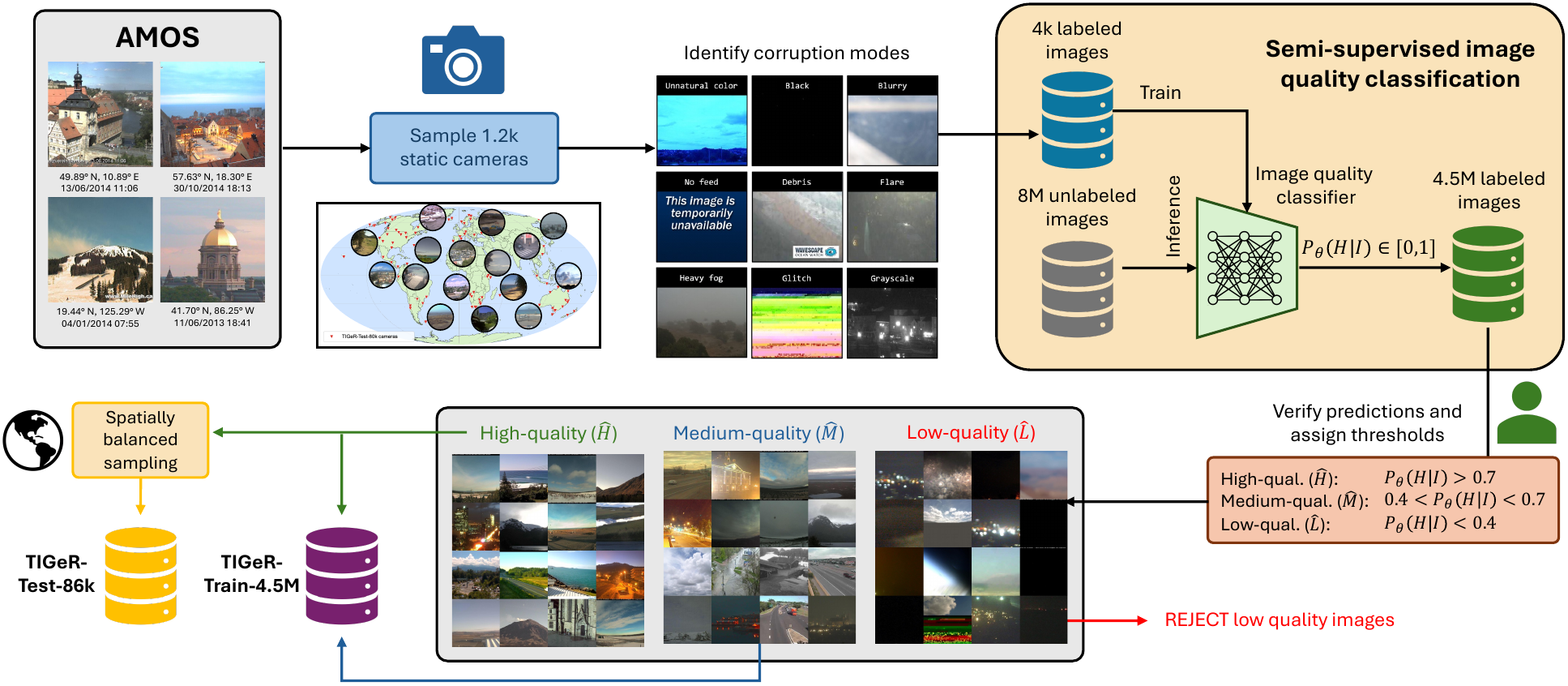}
\end{center}
\vspace{-1em}
\caption{
\textbf{Benchmark dataset curation pipeline.} Starting from the AMOS corpus, we randomly sample 1,255 static cameras with broad global coverage and identify diverse corruption modes. We manually label 4,000 images for quality, train a semi-supervised image quality classifier on frozen visual features, and apply it to 8M unlabeled frames to predict a quality score $P(H|I)$. Using thresholds $T_H{=}0.7$ and $T_L{=}0.4$, images are partitioned into high-, medium-, and low-quality subsets, and low-quality frames are discarded. High-quality images are then used to construct a geographically balanced test set by sampling cameras across $10^\circ \times 10^\circ$ latitude-longitude bins and retaining only cameras with at least 500 frames spanning a year, while the remaining high- and medium-quality images form the training set. The resulting benchmark contains 4.5M training images and 86k test images, with no camera overlap between splits.
}
\label{fig:data_pipeline}
\end{figure*}

Constructing a dataset suitable for time-aware geolocalization presents significant challenges. One potential dataset is AMOS \cite{jacobs07amos, jacobs09webcamgis}, which  provides a vast collection of static outdoor imagery distributed across diverse regions, months, and times-of-day, however, its raw form is highly unsuitable for direct use. The dataset suffers from pervasive noise and corruption, stemming from both hardware limitations and environmental conditions. Many cameras frequently generate unusable frames due to broken hardware or missing camera feeds. Low-cost sensors introduce additive Gaussian-like noise and compression artifacts, while night time images often collapse into near-black frames with severely limited dynamic range. Other failure modes include misconfigured sensors producing banding or dead pixels, placeholders or stale frames with uniform content, and weather-induced distortions (rain, snow, fog) that obscure critical scene details. Without careful filtering, these corruptions overwhelm useful signal, making curation a central challenge. We list all the corruption modes and describe them in detail in the supplementary section~\ref{sup:sec_corruption}.

To overcome these challenges, we design a multi-stage curation pipeline (Figure~\ref{fig:data_pipeline}), that systematically filters noise while preserving both geographic and temporal diversity. To ensure global coverage, we begin by randomly sampling 1,255 cameras from the AMOS dataset across both hemispheres and multiple continents. To build a supervised filtering model, we manually annotate 4,000 samples into high- and low-quality categories, capturing the diverse corruption modes described above. Formally, if an image $I$ has any form of corruption, we assign it to the low-quality set $L$ with a label $q=0$; otherwise it is placed in the high-quality set $H$ with $q=1$. 

Using these seed sets, we train a lightweight classifier on frozen CLIP \cite{clip} and DINOv2 \cite{oquab2023dinov2} visual features, with a linear probe optimized for binary quality prediction. The output of the model is a confidence score $p(q=H|I) \in [0,1]$, indicating the probability that an image is of high-quality. We then evaluate the model's performance on a hold-out set of 400 images resulting in a binary classification accuracy of 91\%. Based on the cumulative score distribution to select two thresholds $T_H=0.7$ and $T_L=0.4$ which we use to group the predictions into high- ($\hat H$), medium- ($\hat M$), and low-quality ($\hat L$) sets as follows:
\begin{equation}
    I \in \begin{cases}
        \hat H, & \mathrm{if} \quad P(H|I) \geq T_H \\
        \hat M, & \mathrm{if} \quad T_L \leq P(H|I) < T_H  \\
        \hat L, & \mathrm{if} \quad P(H|I) < T_L
    \end{cases}.
\end{equation}

The images assigned to the low-quality set $\hat L$ are removed from the dataset, since excessive noise can obscure the geo-temporal signals needed to train and evaluate our model. In contrast, the high-quality images in $\hat H$ are well suited for building the test set, as they typically exhibit good illumination, mild weather conditions, and reliable camera feeds. To construct a representative benchmark, we seek a test set with a balanced geographic distribution across both hemispheres. This is particularly important for tasks involving time-of-year, since seasonal appearance is strongly correlated with calendar month, while seasons are inverted between the Northern and Southern Hemispheres.

Beyond balancing the number of test images across hemispheres, we also aim to avoid over-representing regions with denser camera coverage. To this end, we partition the globe into $10^\circ \times 10^\circ$ latitude-longitude bins and select cameras that are approximately evenly distributed across them. From these candidate cameras, we retain only those with at least 500 frames spanning a full year, ensuring coverage across multiple months and hours at each location. All remaining cameras from the high- and medium-quality subsets are assigned to the training set. Importantly, the training and test splits share no cameras, allowing our evaluation to better measure the model’s ability to generalize to unseen locations.

\begin{figure}[t]
\begin{center}
\includegraphics[width=0.9\linewidth]{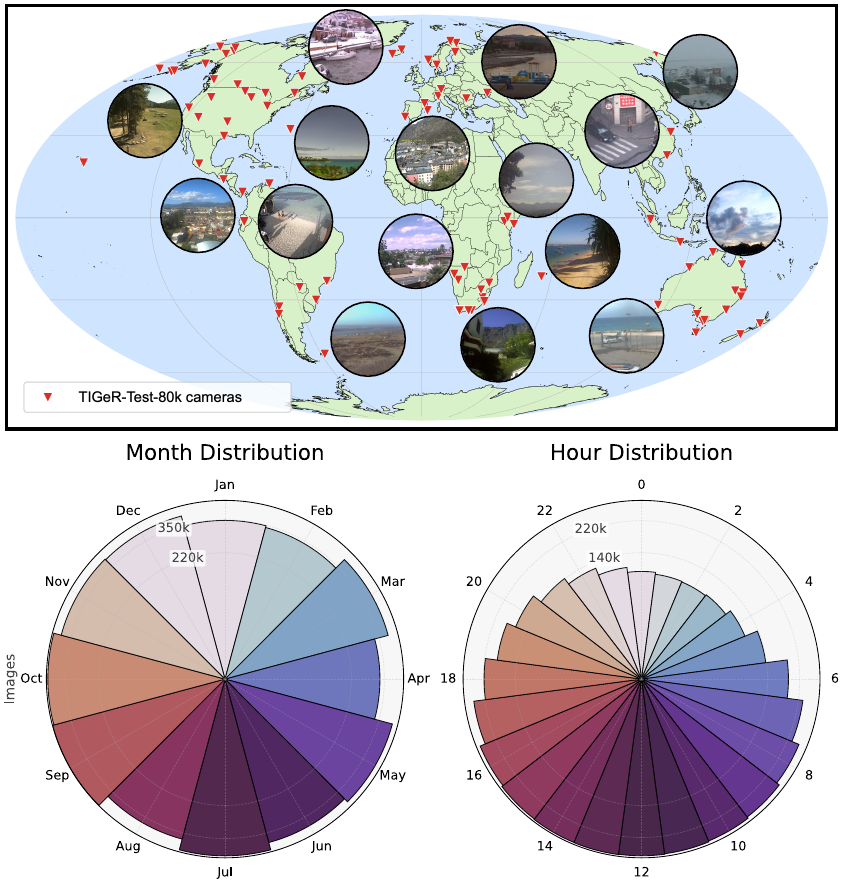}
\end{center}
\vspace{-1em}
\caption{\textbf{Dataset statistics and distributions.}
\textbf{Top}: Geographical distribution of camera locations across the world in the proposed dataset, showing wide coverage across multiple continents.
\textbf{Left}: Time-of-year (month) distribution of the captured images.
\textbf{Right}: Time-of-day (hour) distribution of the captured images.
}
\label{fig:dataset_dist}
\end{figure}

After all the filtering stages, the curated dataset contains $4.5M$ training images and $86k$ test images. Figure~\ref{fig:dataset_dist} illustrates the resulting geographic and temporal distributions. This pipeline transforms the noisy AMOS corpus into a clean, balanced, and diverse benchmark, enabling robust training and reliable evaluation for downstream geo-localization and time prediction tasks.

%% file: sec/5_exp.tex
\section{Benchmarking}

\begin{table*}
    \centering
    \setlength{\tabcolsep}{2pt}
    \renewcommand{\arraystretch}{1.0}
    \caption{\textbf{Geo-time Aware Image Retrieval on images from unseen locations}. Given a query image ($I$) and target time ($t$), the goal is to retrieve an image captured at the \textit{same geographic location} as the query but at the \textit{specified target time}. Our model significantly outperforms prior approaches, demonstrating strong geo-temporal alignment and robustness to temporal variation.  *indicates methods we replicate, closely adhering to protocols outlined by prior work
    }
    \label{tab:sota}
    \begin{tabular}{lccccccc} 
        \toprule
        \multirow{2}{*}{\textbf{Method}} & \multirow{2}{*}{\textbf{Retrieval}} & \multicolumn{3}{c}{\textbf{TIGeR-test-86k}} & \multicolumn{3}{c}{\textbf{CVT}} \\
        \cmidrule(lr){3-5} \cmidrule(lr){6-8}  
        &  &\textbf{R@1 (\%)} & \textbf{R@5 (\%)} & \textbf{R@10 (\%)} & \textbf{R@1 (\%)} & \textbf{R@5 (\%)} & \textbf{R@10 (\%)} \\
        \midrule
        Zhai et al.~\cite{zhai2019learning}* & $It \rightarrow I$ & 1.97 & 7.05 & 11.23 & 0.45 & 1.59 & 2.38 \\
        Zhai et al. w/ CLIP~\cite{zhai2019learning}* & $It \rightarrow I$ & 2.60 & 8.80 & 13.70 & 2.95 & 8.76 & 13.31 \\
        Time-Loc~\cite{Shatwell_2025_ICCV}* & $It \rightarrow I$ & 0.38 & 1.73 & 3.19 & 15.23 & 19.28 & 21.18 \\
        GT-Loc~\cite{Shatwell_2025_ICCV}* & $It \rightarrow I$ & 0.34 & 1.56 & 2.94 & \textbf{16.45} & 20.44 & 22.59 \\
        \midrule
        \textbf{TIGeR (Ours)} & $It \rightarrow I$ & \textbf{3.51} & \textbf{23.30} & \textbf{37.51} & 14.55 & \textbf{24.46} & \textbf{29.98} \\
        \bottomrule
    \end{tabular}
    \vspace{-1em}
\end{table*}

We evaluate \tiger on three retrieval tasks to demonstrate its ability to learn strong and generalizable geo-temporal representations: geo-time aware image retrieval, time-of-capture prediction and geo-localization.

\subsection{Geo-time Aware Image Retrieval}

This task aims to retrieve, from a global gallery, the image that best matches a query in terms of both location and time. Prior work~\cite{Shatwell_2025_ICCV} studied an oracle variant where the ground-truth GPS location $l^Q$ is given, and the query is formed by fusing the location and target time $t^Q$ embeddings. Here we consider a more challenging and realistic setting: given a query image $I^Q$ and target time $t^Q$, the goal is to retrieve an image $\{I_i^G\}_{i=1}^N$ captured at the same physical location as $I^Q$ and at time $t^Q$. In this case, the model must infer and preserve location identity directly from the image, rather than relying on explicit GPS.

We address this problem by encoding the image and time with \tiger and fusing their embeddings into a unified geo-temporal representation. We precompute gallery embeddings $\bar{\mathbf{v}}_i^G = f(I_i^G)$ and, at inference, obtain the fused query representation $\bar{\mathbf{vt}}^Q = f(I^Q, t^Q)$. Retrieval is then performed by maximizing cosine similarity:
\begin{equation}
    (I^Q)^* = \argmax_{I^G} \big( f(I^Q, t^Q)^\top f(I^G) \big).
    \label{eq:optim-gtret}
\end{equation}

We evaluate on two complementary benchmarks: our \textsc{TIGeR-test-86k} split and a balanced partition of CVT~\cite{salem2020dynamic} across hemispheres. \textsc{TIGeR-test-86k} is built from webcams with multiple observations of the same camera across the year, enabling fine-grained geo-temporal retrieval at fixed locations but varying times. CVT instead consists of social media images at arbitrary locations and times, without repeated cameras; here, retrieval is considered correct if the result lies within $125$\,km of the query location and matches the specified time.

We compare \tiger against two main baselines. The first is the triple-encoder classifier of Zhai et al.~\cite{zhai2019learning}, originally proposed for joint geo-location and time-of-capture classification. In addition to the original InceptionV2~\cite{szegedy2016rethinking} backbone, we also evaluate a stronger variant with a CLIP ViT-L/14~\cite{clip} backbone for fair comparison. Following~\cite{Shatwell_2025_ICCV}, we adapt this model to retrieval by first predicting the query geo-location class $\hat{l}^Q$, then computing location and time posteriors $P(l \mid I_i^G)$ and $P(t \mid I_i^G)$ for each gallery image and defining
\[
\mathrm{score}(I_i^G) = P(l = \hat{l}^Q \mid I_i^G)\, P(t = t^Q \mid I_i^G),
\]
which is used to rank the gallery.

The second baseline is GT-Loc~\cite{Shatwell_2025_ICCV}, which maps locations, images, and times into a shared feature space but lacks \tiger's multimodal transformer. Here we encode image and time independently and perform late fusion by averaging their embeddings, which cannot explicitly model higher-order interactions between geo-location and time.

\begin{figure}[t]
\begin{center}
\includegraphics[width=\linewidth]{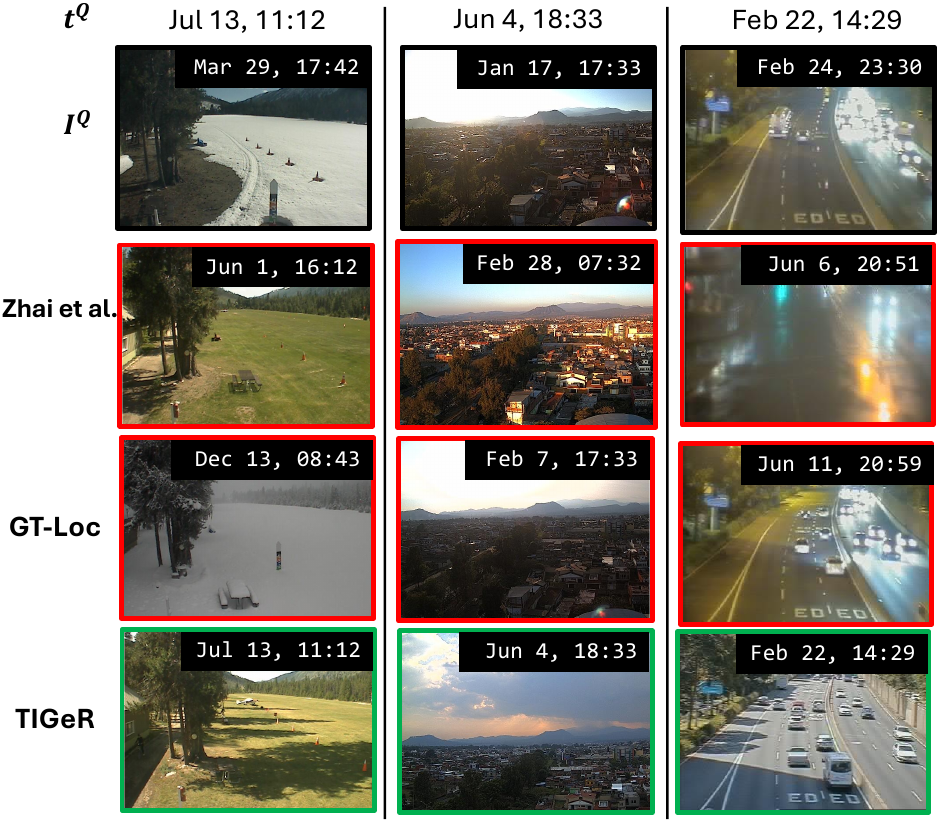}
\end{center}
\caption{\textbf{Qualitative results on geo-time aware image retrieval with images from unseen locations}. Given a query image $I^Q$ and a target time $t^Q$ (shown above each column), the goal is to retrieve an image captured at the \textit{same location} as the query but at the specified target time. For visualization, the actual time-of-capture for each image is overlaid on top of it. 
Each row shows top-1 retrieval results from different methods. While prior approaches (Zhai et al.~\cite{zhai2019learning}, GT-Loc~\cite{Shatwell_2025_ICCV}) often retrieve images captured at incorrect times, our \textsc{\tiger} model accurately retrieves images corresponding to the target time while maintaining location consistency. Best viewed in color. 
}
\label{fig:qualitative}
\end{figure}

Table~\ref{tab:sota} reports quantitative results. \tiger consistently outperforms all baselines, achieving higher recall at most ranks on both datasets. This indicates that the learned representations are well aligned across visual, temporal, and geo-location modalities, and that the fused embeddings successfully encode both location and time information rather than over-emphasizing purely semantic visual similarity. Qualitative examples comparing \tiger and the baselines are shown in Figure~\ref{fig:qualitative}. We show additional examples on the supplementary section~\ref{sup:qualitative}.

\subsection{Time Prediction and Geo-localization}

In this task, the objective is to estimate both the geo-location and time-of-capture of a query image by comparing its embedding against dedicated galleries of location and time embeddings, respectively, and selecting the most compatible entries according to cosine similarity.

Let $I^{Q}$ be a query image and let $\{x_i^{G}\}_{i=1}^{N_G}$ denote a gallery of targets from a modality $x$ (either GPS coordinates or timestamps). Our model produces an image embedding $\bar{\mathbf{v}}^{Q}\!\in\!\mathbb{R}^{d}$ and a gallery embedding $\mathbf{x}^{G}_i\!\in\!\mathbb{R}^{d}$ for each gallery item. In addition, a modality-specific classifier predicts a probability vector $\hat{\mathbf{y}}$ over $B$ coarse classes (e.g., HEALPix bins for GPS or time bins for timestamps). We write $b(x_i^{G})\!\in\!\{1,\dots,B\}$ for the class index associated with gallery item $x_i^{G}$.

We combine a continuous retrieval signal with a discrete prior. Treat the cosine similarity as a temperature-scaled log-likelihood of a match, and the classifier output as a prior over classes. The resulting posterior-like score for each gallery item $x_i^{G}$ is
\begin{equation}
    \label{eq:poe-score}
    \mathrm{score}(x_i^{G})
    \;=\;
    \frac{(\bar{\mathbf{v}}^{Q})^{\top}\mathbf{x}^{G}_i}{\psi}
    \;+\;
    \beta(I^{Q})\,\log P\!\left(b(x_i^{G}) \mid I^{Q}\right),
\end{equation}
where $\psi\!>\!0$ is a temperature parameter and $\beta(I^{Q})\!\ge\!0$ controls the influence of the classifier prior for the query.

A fixed $\beta$ can help on easy (confident) queries and hurt on hard (uncertain) ones. We therefore make $\beta$ \emph{query-adaptive} using the classifier entropy:
\begin{equation}
    \label{eq:beta-entropy}
    \beta(I^{Q})
    \;=\;
    \beta_{\max}\!\left(
    1 - \frac{H(\hat{\mathbf{y}})}{\log B}
    \right),
\end{equation}
where $H(\hat{\mathbf{y}})$ is the Shannon entropy and $\log B$ the maximum entropy for $B$ classes. Thus, $\beta(I^{Q})\!\in[0,\beta_{\max}]$: it approaches $\beta_{\max}$ when the classifier is confident (low entropy) and decreases toward $0$ when it is uncertain (high entropy).

Equation~\eqref{eq:poe-score} allows the retrieval term to \emph{pull} fine-grained neighbors while the prior \emph{guides} the search toward plausible classes. The entropy-adaptive weighting in \eqref{eq:beta-entropy} ensures the prior has strong influence only when it is informative, mitigating boundary effects and domain shift.

In the time prediction task, rather than conditioning onlt on the image embedding $\bar{\mathbf{v}}^{Q}$, we can optionally incorporate the known location as auxiliary input. In this setting, we use the fused embedding $\bar{\mathbf{vl}}^{Q}=f(I^{Q},G^{Q})$ as the query representation and compute cosine similarity against gallery time embeddings. This variant often improves temporal prediction accuracy by constraining retrieval to locations consistent with the provided GPS context.

Tables~\ref{tab:time-pred} and~\ref{tab:geolocalization} summarize the performance of our method on time prediction and geo-localization, respectively. Overall, \tiger achieves state-of-the-art results on both tasks. For time-of-capture prediction, \tiger consistently outperforms prior methods on both month and hour estimation. Moreover, when geo-location is provided as an auxiliary input and fused with the image representation, performance improves substantially. This behavior is expected, as the visual appearance of time is strongly coupled with latitude and season (e.g., higher latitudes exhibit more pronounced seasonal variation than regions near the Equator).

For geo-localization, \tiger also attains competitive performance compared to strong baselines. However, in this setting, using time as an auxiliary signal does not always lead to further gains. Unlike location, time-of-capture is a comparatively weak and ambiguous cue for disambiguating places: many geographically distant locations can share similar illumination and seasonal appearance at a given time, and the time predictions themselves may be noisy. Incorporating this uncertain temporal signal can therefore introduce additional variance or bias the model toward spurious correlations, which sometimes offsets the benefits of joint time–location reasoning for geo-localization.

\begin{table}[t]
\centering
\setlength{\aboverulesep}{0pt}
\setlength{\belowrulesep}{0pt}
\caption{\textbf{Time prediction} on unseen cameras. *indicates methods we replicate, closely adhering to protocols outlined by prior work.}
\label{tab:time-pred}
\begingroup
\setlength{\tabcolsep}{1pt}
\resizebox{\linewidth}{!}{
\begin{tabular}{lccccc} 
\toprule
\multirow{3}{*}{\textbf{Method}} & \multirow{3}{*}{\textbf{Retrieval}}
  & \multicolumn{2}{c}{\textbf{TIGeR-test-86k}}
  & \multicolumn{2}{c}{\textbf{CVT-test}} \\
\cmidrule(lr){3-4} \cmidrule(lr){5-6}
                & 
  & \textbf{ToY} & \textbf{ToD} 
  & \textbf{ToY} & \textbf{ToD}  \\
                & 
  & \textbf{Error}$\downarrow$ & \textbf{Error}$\downarrow$
  & \textbf{Error}$\downarrow$ & \textbf{Error}$\downarrow$  \\
\midrule

Zhai et al.~\cite{zhai2019learning}*                & $I \rightarrow t$ & 68.38 & 3.97 & 87.37 & 3.28  \\
Zhai et al. w/ CLIP ~\cite{zhai2019learning}*        & $I \rightarrow t$ & 57.51 & 3.22 & 68.95 & 2.8  \\
Time-Loc~\cite{Shatwell_2025_ICCV}*   & $I \rightarrow t$ & 74.87 & 4.02 & 65.10 & 2.86 \\
GT-Loc~\cite{Shatwell_2025_ICCV}*    & $I \rightarrow t$  & 74.58 & 3.52 & 78.95 & \textbf{2.68} \\

\midrule

\textbf{TIGeR (Ours)}    & $I \rightarrow t$  & \textbf{51.49} & \textbf{3.13} & \textbf{62.88} & 2.73 \\

\midrule
Zhai et al.~\cite{zhai2019learning}*               & $Il \rightarrow t$ & 66.36 & 3.88 & 73.15 & 3.15  \\
Salem et al.~\cite{salem2022timestamp}*               & $Il \rightarrow t$ & 74.29 & 3.57 & 75.48 & 3.1  \\
Zhai et al. w/ CLIP~\cite{zhai2019learning}*        & $Il \rightarrow t$ & 55.15 & 3.19 & 60.42 & 2.77  \\
Salem et al. w/ CLIP~\cite{salem2022timestamp}*       & $Il \rightarrow t$ & 55.4 & 3.24 & 64.05 & 2.71  \\

\midrule

\textbf{TIGeR (Ours)}    & $Il \rightarrow t$ & \textbf{48.86} & \textbf{3.06} & \textbf{47.00} & \textbf{2.35} \\

\bottomrule
\end{tabular}}
\endgroup
\end{table}

\begin{table}
    \centering
    \caption{\textbf{Geo-localization accuracy}. *indicates methods we replicate, closely adhering to protocols outlined by prior work.}
    \label{tab:geolocalization}
    \resizebox{\columnwidth}{!}{%
    \begin{tabular}{lccccc}
        \toprule
        \multirow{2}{*}{\textbf{Method}} &
        \multirow{2}{*}{\textbf{Retrieval}} &
        \multicolumn{2}{c}{\textbf{TIGeR-test-86k}} &
        \multicolumn{2}{c}{\textbf{CVT-test}} \\
        \cmidrule(lr){3-4}\cmidrule(lr){5-6}
        & & \textbf{200 km} & \textbf{750 km} &
            \textbf{200 km} & \textbf{750 km} \\
        \midrule
        Zhai et al.~\cite{zhai2019learning}*          & $I \rightarrow l$  & 22.05 & 46.20 &  3.50 & 10.50 \\
        Zhai et al. w/ CLIP~\cite{zhai2019learning}*   & $I \rightarrow l$  & 28.58 & 64.85 & 23.13 & 57.74 \\
        GeoCLIP~\cite{vivanco2024geoclip}              & $I \rightarrow l$  & 21.58 & 37.58 & 43.62 & 60.77 \\
        GT-Loc~\cite{Shatwell_2025_ICCV}*               & $I \rightarrow l$  & 21.07 & 35.93 & 42.63 & 59.49 \\
        \midrule
        \textbf{TIGeR (Ours)}        & $I \rightarrow l$  & \textbf{48.63} & \textbf{65.61} & \textbf{51.90} & \textbf{66.53} \\
        \midrule
        Zhai et al.~\cite{zhai2019learning}*          & $It \rightarrow l$ & 21.99 & 46.62 &  4.46 & 13.41 \\
        Zhai et al. w/ CLIP~\cite{zhai2019learning}*   & $It \rightarrow l$ & 28.83 & 65.16 & 23.82 & 58.97 \\
        \midrule
        \textbf{TIGeR (Ours)}        & $It \rightarrow l$ & \textbf{48.34} & \textbf{64.51} & \textbf{53.40} & \textbf{67.15} \\
        \bottomrule
    \end{tabular}%
    }
\end{table}

\begin{table}[h]
    \centering
    \caption{Cumulative ablation study on the components of \tiger.}
    \label{tab:ablations-components}
    \resizebox{\columnwidth}{!}{%
    \begin{tabular}{lccc} 
        \toprule
        
        \multirow{2}{*}{\textbf{Setting}} &
        \multicolumn{1}{c}{\textbf{Month}} &
        \multicolumn{1}{c}{\textbf{Hour}} &
        \multicolumn{1}{c}{\textbf{Geo-location}} \\
        & \multicolumn{1}{c}{\textbf{Error}} &
          \multicolumn{1}{c}{\textbf{Error}} &
          \multicolumn{1}{c}{\textbf{Error (km)}} \\
        \midrule
        Base model                      & 76.77 & 3.10 & 668.97 \\
        + multi-modal transformer       & 57.58 & 2.98 & 238.58 \\
        + Classification heads          & 57.20 & 2.95 & 234.53 \\
        + Entropy adaptive reranking    & 57.20 & 2.93 & 203.93 \\
        \bottomrule
    \end{tabular}%
    }
\end{table}

\subsection{Ablation Analysis}

We conduct an ablation study to quantify the contribution of each component of \tiger on geo-localization and time-of-capture prediction. For both tasks, we report errors averaged over \textsc{TIGeR-test-86k} and CVT-test. For geo-localization, we use mean geodetic error (in kilometers) rather than recall at distance thresholds, as this yields a single, directly comparable metric.

Table~\ref{tab:ablations-components} summarizes the results. We start from a \emph{base model} that includes image, location, and time encoders trained with only two contrastive losses, analogous to GT-Loc~\cite{Shatwell_2025_ICCV}. Adding the \emph{multi-modal transformer} together with the five cross-modal contrastive losses from Eq.~\ref{eq:contrastive-losses} yields the largest performance gain across both temporal and geo-location errors, highlighting the importance of joint geo-temporal modeling. Next, we introduce \emph{classification heads} as auxiliary supervision during training, which provides a modest additional improvement. Finally, we leverage these heads at test time through our \emph{entropy-adaptive reranking} scheme, which uses the classifier predictions as a confidence-weighted prior. This step produces a substantial reduction in geo-localization error, resulting in the full \tiger model. Additional ablations and analyses are provided in the supplementary section~\ref{sup:ablations}.

%% file: sec/6_conclusion.tex
\section{Conclusion}

We introduced \tiger, a unified framework for jointly modeling images, geo-location, and time, and formalized the task of \emph{geo-time aware image retrieval}, where the goal is to retrieve an image captured at the same physical location as a query but at a specified target time. To enable rigorous evaluation, we curated a large-scale benchmark with globally distributed, temporally diverse, and carefully filtered imagery. \tiger combines modality-specific encoders with a multi-modal transformer to map all modalities into a shared geo-temporal embedding space, and is trained with contrastive objectives and entropy-adaptive reranking to capture fine-grained relationships between where and when a scene is observed. Extensive experiments show that \tiger achieves state-of-the-art performance across geo-localization, time-of-capture prediction, and geo-time aware image retrieval.

%% file: sec/7_acknowledgement.tex
\section*{Acknowledgements}

Supported by Intelligence Advanced Research Projects Activity (IARPA) via Department of Interior/Interior Business Center (DOI/IBC) contract number 140D0423C0074. The U.S. Government is authorized to reproduce and distribute reprints for Governmental purposes notwithstanding any copyright annotation thereon. Disclaimer: The views and conclusions contained herein are those of the authors and should not be interpreted as necessarily representing the official policies or endorsements, either expressed or implied, of IARPA, DOI/IBC, or the U.S. Government.

%% file: sec/X_suppl.tex
\clearpage
\setcounter{page}{1}
\maketitlesupplementary

\setcounter{section}{0}%
\renewcommand{\thesection}{\Alph{section}}

\noindent We organize the supplementary as follows: 

\begin{itemize}

    \item Section~\ref{sup:sec_corruption}: Corruption modes in AMOS
    \item Section~\ref{sup:qualitative}: Qualitative results
    \item Section~\ref{sup:ablations}: Ablation analysis
    \item Section~\ref{sup:compositional}: Compositional image retrieval
    \item Section~\ref{sup:gt-cond-img-ret}: Geo-temporal image retrieval analysis
    \item Section~\ref{sup:time-pred}: Time prediction analysis
    \item Section~\ref{sup:training-details}: Training details
    \item Section~\ref{sup:embedding}: Embedding space visualization

\end{itemize}

\section{Corruption modes in AMOS}
\label{sup:sec_corruption}

\setlength{\columnsep}{6pt}%
\begin{wrapfigure}[9]{r}{0.5\linewidth}
\vspace{-16pt}
  \centering
  \includegraphics[width=\linewidth, trim=26 14 0 0, clip]{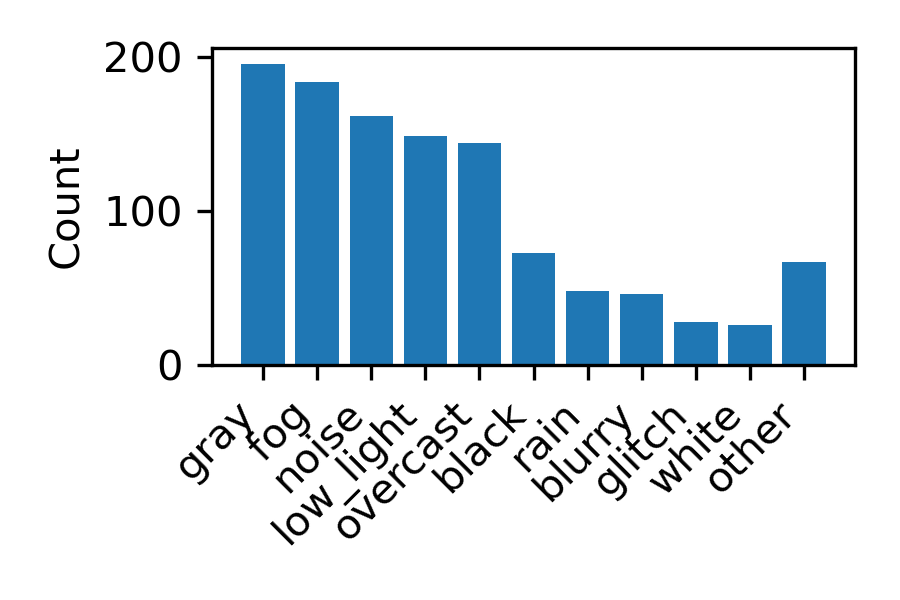}
    \vspace{-7mm}
    \captionsetup{font=small}
    \caption{\bf{Distribution of corruption modes in AMOS.}}
  \label{fig:corr-count}
\end{wrapfigure}

\begin{figure}[t]
\begin{center}
\includegraphics[width=0.95\linewidth]{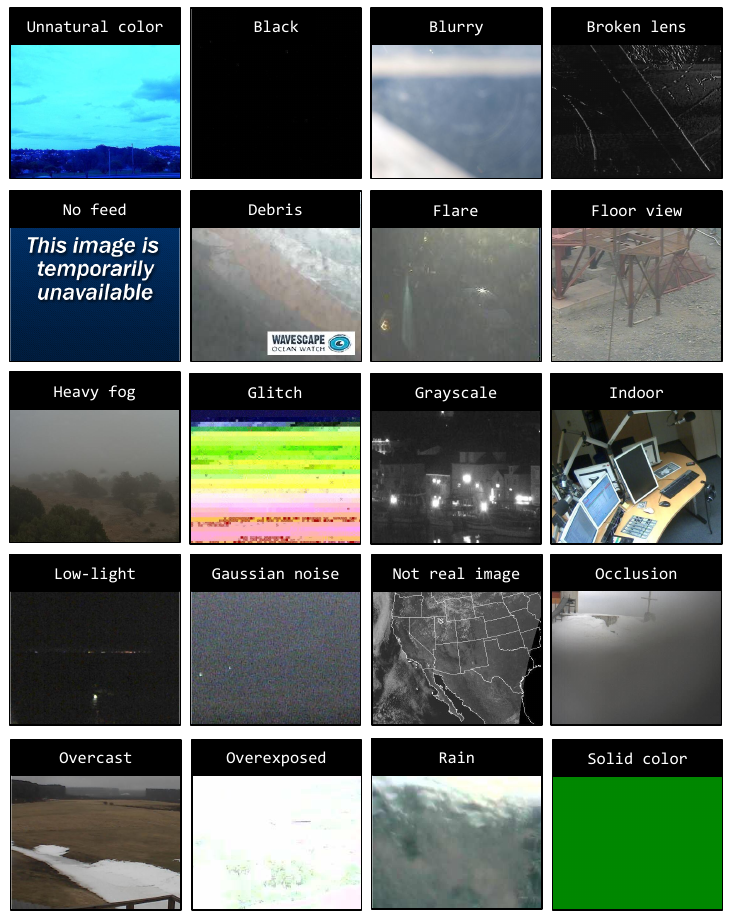}
\end{center}
\vspace{-1em}
\caption{\textbf{Corruption modes in the AMOS dataset.}
Examples of typical failure modes in raw AMOS imagery, including sensor or hardware issues (black frames, dead pixels, broken lenses, ``no feed'' screens), challenging environmental conditions (fog, rain, nighttime), imaging and compression artifacts (blur, glare, noise, color shifts), and content or framing errors (indoor views, floor shots, overlays, or static graphics). 
These corruptions are frequent and often persistent over time, underscoring the need for the filtering pipeline described in Section~\ref{sec:data_curation}.}
\label{fig:corruptions}
\end{figure}

\begin{figure}[t]
\begin{center}
\includegraphics[width=0.98\linewidth]{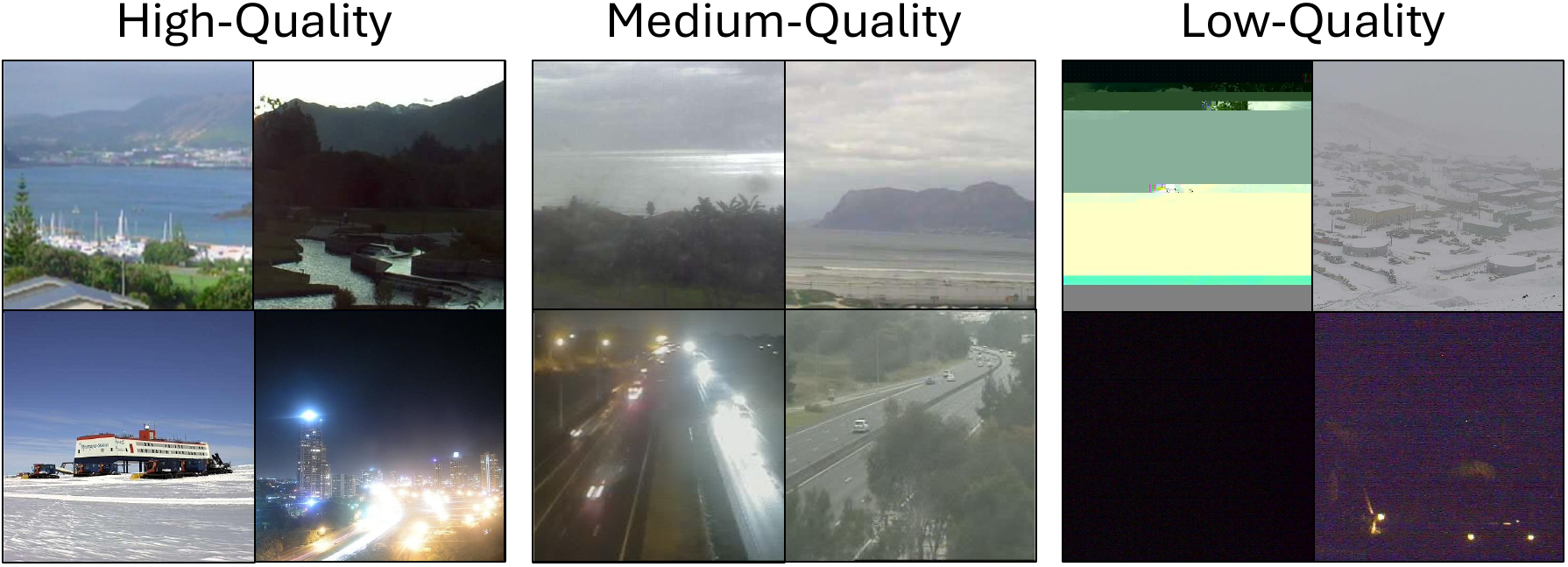}
\end{center}
\vspace{-1em}
\caption{Sample images that were classified as high, medium or low-quality by our semi-supervised data curation pipeline.}
\label{fig:quality_examples}
\end{figure}

Figure~\ref{fig:corr-count} presents the distribution of corruption modes observed in the manually annotated samples from the original AMOS dataset. Complementary qualitative examples are shown in Figure~\ref{fig:corruptions}, highlighting the wide variability and visual characteristics of these corruptions. The observed artifacts stem from a combination of environmental conditions, sensor and hardware degradation, and network-related issues. As a result, the dataset cannot be reliably used in its raw form and requires substantial filtering prior to downstream applications.
We categorize these failure modes into four broad groups: (i) \textbf{Sensor or hardware failures}, including black frames, broken lenses, dead pixels, and “no feed” placeholders; (ii) \textbf{Environmental conditions}, such as fog, rain, overcast skies, or low-light nighttime scenes; (iii) \textbf{Imaging and compression artifacts}, including blur, flare, Gaussian noise, glitches, or unnatural color balance; and (iv) \textbf{Content or framing errors}, such as indoor captures, floor views, or non-visual placeholders (e.g., weather maps, static logos).
These corruptions can dominate the raw corpus, with many cameras producing long sequences of invalid or heavily degraded frames. To construct a reliable benchmark, we developed a semi-supervised filtering pipeline (Section~\ref{sec:data_curation}, Figure~\ref{fig:data_pipeline} in the main paper) that automatically detects and removes such samples while preserving geographic and temporal diversity. We show additional examples of images that were ultimately classified as high, medium and low-quality in Figure~\ref{fig:quality_examples}.

\section{Qualitative results}
\label{sup:qualitative}

\begin{figure*}
\begin{center}
\includegraphics[width=0.9\linewidth]{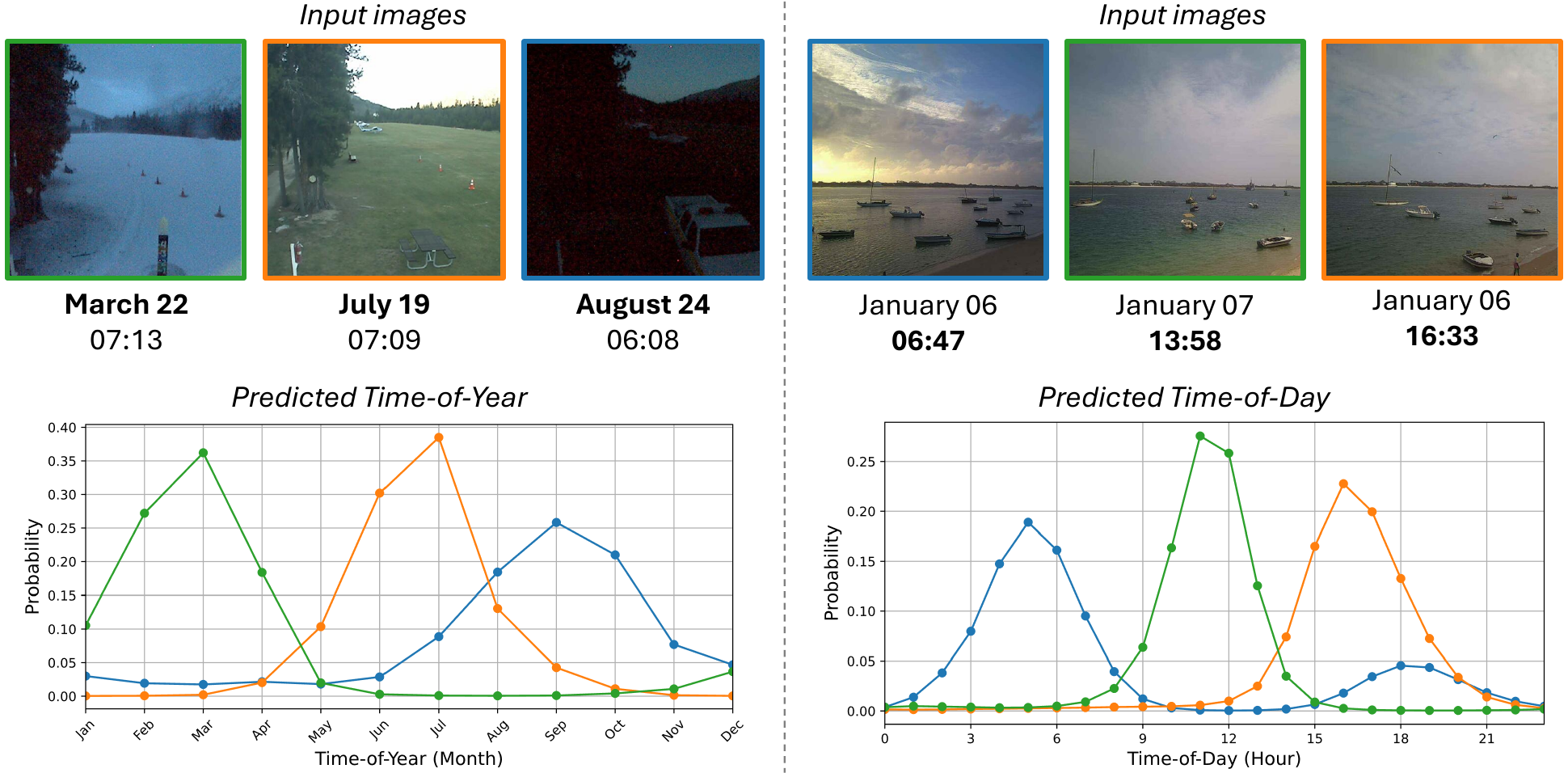}
\end{center}
\vspace{-1em}
\caption{\textbf{Time prediction probabilities}. For an input image $I^Q$, we compute cosine similarity between it's visual embedding $\bar{\textbf{v}}^Q$ and a gallery of time embeddings $\bar{\textbf{t}}^G$, sampled at 1-hour and 1-month intervals. The similarity vector is normalized using the softmax operator, resulting in a predicted time probability distribution.
\textbf{Left}: Comparison of the predicted time-of-year probabilities for images at roughly the same time-of-day, but different time-of-years. \textbf{Right}: Comparison of the predicted time-of-day probabilities for images at roughly the same time-of-year, but different time-of-days.}
\label{fig:time-pred-dist}
\end{figure*}

\noindent \textbf{Time prediction.} Figure~\ref{fig:time-pred-dist} illustrates how \tiger\ models time from a single image. For each query image, we compute cosine similarities between its visual embedding and a gallery of time embeddings, then aggregate the scores into 1-month (or 1-hour) bins and apply a softmax to obtain a probability distribution over time-of-year (ToY) or time-of-day (ToD). On the left, we show three queries captured around the same ToD (between 6--7 AM) but in different months (March, July, and August); the predicted distributions are defined over months and peak near the ground-truth ToY, with most mass concentrated in neighboring bins. On the right, the queries all come from January but at different hours; here the distributions are defined over hours and again peak near the ground-truth ToD. These results indicate that \tiger\ produces sharp, well-localized temporal distributions that accurately reflect both ToY and ToD.

\begin{figure}[h]
\begin{center}
\includegraphics[width=0.9\linewidth]{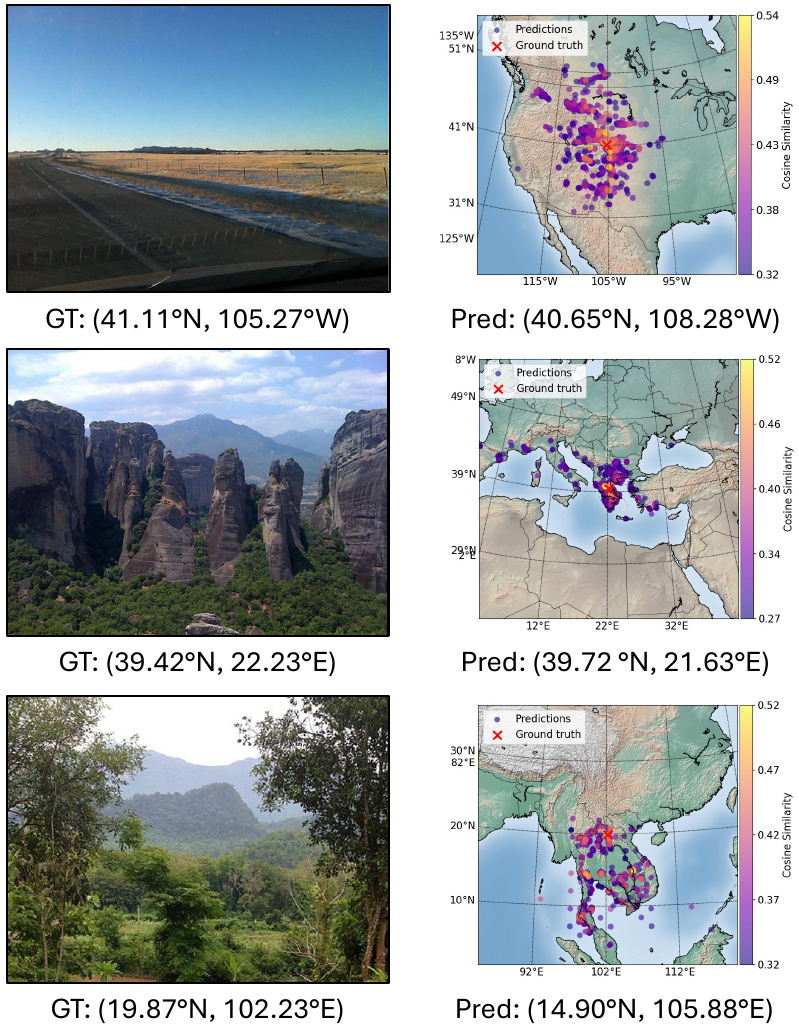}
\end{center}
\vspace{-1em}
\caption{\textbf{Qualitative geo-localization examples on CVT.} Each row shows a query image (left) annotated with its ground-truth GPS coordinate, and the corresponding map of retrieved locations (right). We plot the top-1k candidate GPS coordinates predicted by \tiger, color-coded by cosine similarity with the image embedding (lighter colors indicate higher similarity). The top-1 prediction is highlighted in yellow and the ground truth is marked with a red $\times$. In all three cases, high-similarity predictions concentrate near the ground truth, indicating accurate localization across diverse scenes and regions.}
\label{fig:qual-geo}
\end{figure}

\noindent \textbf{Geo-localization.} Figure~\ref{fig:qual-geo} shows how \tiger\ localizes three CVT test images taken in geographically diverse regions. For each query image, we display the corresponding distribution of retrieved GPS candidates, highlighting both the predicted and ground-truth locations to illustrate the sharpness and accuracy of the model’s geo-localization.

\section{Ablations}
\label{sup:ablations}
In this section, we provide additional ablation analysis of the design choices in \tiger.  
Our ablations evaluate, (i) the effect of the shared multimodal fusion transformer, (ii) the effect of the pretrained image backbone, and (iii) The effect of geo-location and temporal thresholds on the image retrieval tasks.

\subsection{Effect of the shared multimodal fusion transformer}

To assess the benefit of the shared multimodal fusion transformer, we compare our full model with a late-fusion baseline. In this baseline, three independent transformers are placed on top of the image, location, and time encoders, and the resulting modality-specific embeddings are only combined at the end by simple averaging (late fusion). In contrast, our model feeds all modalities into a single transformer, enabling early cross-modal interaction.

Table~\ref{tab:ablations-fusion} reports the average performance of both variants on tasks that explicitly require combining modalities: time-aware geo-localization ($I t \rightarrow l$), location-aware time prediction ($I l \rightarrow t$), and geo-time-aware image retrieval ($I t \rightarrow I$), averaged over \textsc{TIGeR-test-86k} and CVT. Using the shared fusion transformer consistently improves all metrics: ToY error decreases by 10.5\%, ToD error by 7.43\%, and geo-localization error by 26.76\%, while geo-time-aware image retrieval gains an absolute \mbox{11.91\%} in R@10. These results highlight the importance of early multimodal fusion for accurate geo-temporal reasoning.

\begin{table}
    \centering
    \caption{\textbf{Effect of the shared multimodal fusion transformer.} 
    We compare a late-fusion baseline (separate transformers for each modality, combined by averaging) with our shared transformer that performs early fusion of image, location, and time embeddings. 
    We report time-of-year (ToY) and time-of-day (ToD) errors, geo-location error (km), and geo-time-aware image retrieval R@10, averaged over \textsc{TIGeR-test-86k} and CVT (lower is better for errors, higher is better for R@10). 
    The shared fusion transformer yields consistent gains across all metrics.}
    \label{tab:ablations-fusion}
    
    \resizebox{\columnwidth}{!}{
    \begin{tabular}{ccccc} 
        \toprule
        
        \multicolumn{1}{c}{\textbf{Multimodal Shared}} &
        \multicolumn{1}{c}{\textbf{ToY}} &
        \multicolumn{1}{c}{\textbf{ToD}} &
        \multicolumn{1}{c}{\textbf{Geo-location}} & 
        \multicolumn{1}{c}{\textbf{Geo-time}} \\
        \multicolumn{1}{c}{\textbf{Fusion Transformer}} & 
        \multicolumn{1}{c}{\textbf{Error}} &
          \multicolumn{1}{c}{\textbf{Error}} &
          \multicolumn{1}{c}{\textbf{Error (km)}} & 
          \multicolumn{1}{c}{\textbf{R@10}} \\
        \midrule
        $\times$     & 53.90 & 2.96 & 355.52 & 22.09 \\
        $\checkmark$ & \textbf{48.24} & \textbf{2.74} & \textbf{260.39} & \textbf{34.00} \\
        \bottomrule
    \end{tabular}
    }
\end{table}

\subsection{Effect of the image backbone}

Recent years have seen the emergence of several strong image backbones, each with different strengths depending on the downstream task. To identify the most effective backbone for time prediction, geo-localization, and geo-time-aware image retrieval, we perform an ablation study using four widely adopted vision backbones: CLIP \cite{clip}, DINOv2 \cite{oquab2023dinov2}, Masked Autoencoder (MAE) \cite{he2022masked}, and SigLIP~2 \cite{tschannen2025siglip}. For a fair comparison, all experiments use the large ViT variant of each backbone, with model sizes close to 400M parameters. The results in Table~\ref{tab:ablations-backbone} show that CLIP delivers the strongest overall performance across all evaluated tasks, making it the natural backbone choice for \tiger. Among the remaining alternatives, DINOv2 performs most competitively, but still leads to a 13.69\% increase in ToY error, a 2.05\% increase in ToD error, and a 176.47\% increase in geo-localization error relative to CLIP, and 10.55\% decrease in geo-time-aware image retrieval (R@10).

\begin{table}
    \centering
    \caption{\textbf{Effect of the image backbone.} 
    The pretrained CLIP \cite{clip} ViT backbone results in stronger overall performance compared to DINOv2 \cite{oquab2023dinov2}, Masked Autoencoder \cite{he2022masked}, and SigLIP 2 \cite{tschannen2025siglip}.
    We report time-of-year (ToY) and time-of-day (ToD) errors, geo-location error (km), and geo-time-aware image retrieval R@10, averaged over \textsc{TIGeR-test-86k} and CVT (lower is better for errors, higher is better for R@10).}
    \label{tab:ablations-backbone}
    
    \resizebox{\columnwidth}{!}{
    \begin{tabular}{crrrr} 
        \toprule
        
        \multicolumn{1}{c}{\textbf{Image}} &
        \multicolumn{1}{c}{\textbf{ToY}} &
        \multicolumn{1}{c}{\textbf{ToD}} &
        \multicolumn{1}{c}{\textbf{Geo-location}} & 
        \multicolumn{1}{c}{\textbf{Geo-time}} \\
        \multicolumn{1}{c}{\textbf{Backbone}} & 
        \multicolumn{1}{c}{\textbf{Error}} &
          \multicolumn{1}{c}{\textbf{Error}} &
          \multicolumn{1}{c}{\textbf{Error (km)}} & 
          \multicolumn{1}{c}{\textbf{R@10}} \\
        \midrule
        DINOv2      & 65.02 & 2.99 &  563.39 & 23.20 \\
        MAE         & 73.90 & 3.26 & 4139.82 & 15.05 \\
        SigLIP 2    & 75.88 & 4.60 & 1965.55 & 16.71 \\
        \bf CLIP    & \bf 57.19 & \bf 2.93 &  \bf 203.78 & \bf 33.75 \\
        
        \bottomrule
    \end{tabular}
    }
\end{table}

\subsection{Effect of geo-location and temporal thresholds in geo-time-aware image retrieval}

In the geo-time-aware image retrieval task, a prediction is considered correct if the retrieved image is within a spatial radius $T_{\text{geo}}$ of the query location and within temporal tolerances $T_{\text{ToY}}$ (in days) and $T_{\text{ToD}}$ (in hours). Our goal is therefore not strict instance-level retrieval, but rather to recover an image captured from approximately the same place at the specified target time, which is more realistic given that the gallery may not contain an exact match.

Table~\ref{tab:ablations-threshold} analyzes how sensitive performance is to the choice of these thresholds. For both datasets, we vary each threshold to half and twice its default value and report the resulting recall. On \textsc{TIGeR-test-86k} we adopt a relatively tight spatial threshold of $T_{\text{geo}}=25$ km, since the gallery is known to contain images from exactly the same locations as the queries; on CVT, where this assumption does not hold, we relax the default to $T_{\text{geo}}=125$ km. Across these settings, recall is fairly stable with respect to $T_{\text{geo}}$, while it varies more with the temporal thresholds, indicating that our default choice of 30 days and 1 hour is reasonable yet non-trivial.

\begin{table}
    \centering
    \setlength{\tabcolsep}{2pt}
    \renewcommand{\arraystretch}{1.0}
    \caption{\textbf{Ablations of the image retrieval thresholds} on the geo-time-aware image retrieval task. We change the thresholds used to determine if an was retrieved correctly and observe that the predictions are robust to geo-location distance thresholds $\mathbf{T_{geo}}$, given in kilometers. The results are more sensitive to time prediction thresholds ($\mathbf{T_{ToY}}$, $\mathbf{T_{ToD}}$, given in days and hours respectively), which make intuitive sense, given that our choice of temporal thresholds (15-60 days, 0.5-2 hours) cover a larger portion of the total possible range of times compared to the geo-location threshold (12.5-250 km).}
    \label{tab:ablations-threshold}
    \begin{tabular}{ccccccc} 
        \toprule
        
        \multirow{2}{*}{$\mathbf{T_{geo}}$} & \multirow{2}{*}{$\mathbf{T_{ToY}}$} & \multirow{2}{*}{$\mathbf{T_{ToD}}$} & \multicolumn{3}{c}{\textbf{TIGeR-test-86k}} \\
        \cmidrule(lr){4-6}  
         &  &  & \textbf{R@1 (\%)} & \textbf{R@5 (\%)} & \textbf{R@10 (\%)} \\
        
        \midrule
        
        25.00 & 15 & 0.50 & 1.12 & 9.10 & 16.81 \\
        25.00 & 60 & 2.00 & 11.48 & 48.17 & 64.46 \\
        12.50 & 30 & 1.00 & 3.51 & 23.31 & 37.51 \\
        50.00 & 30 & 1.00 & 3.51 & 23.31 & 37.53 \\
        25.00 & 30 & 1.00 & 3.51 & 23.31 & 37.51 \\

        \midrule

        \multirow{2}{*}{$\mathbf{T_{geo}}$} & \multirow{2}{*}{$\mathbf{T_{ToY}}$} & \multirow{2}{*}{$\mathbf{T_{ToD}}$} & \multicolumn{3}{c}{\textbf{CVT}} \\
        \cmidrule(lr){4-6}
         &  &  & \textbf{R@1 (\%)} & \textbf{R@5 (\%)} & \textbf{R@10 (\%)} \\

        \midrule
        
        125.00 & 15 & 0.50 & 11.85 & 19.33 & 23.49 \\
        125.00 & 60 & 2.00 & 21.79 & 38.54 & 46.86 \\
        62.50 & 30 & 1.00 & 14.25 & 24.53 & 30.26 \\
        250.00 & 30 & 1.00 & 14.55 & 26.46 & 33.24 \\
        125.00 & 30 & 1.00 & 14.55 & 25.51 & 31.69 \\

        \bottomrule
    \end{tabular}
\end{table}

\vspace{-1em}
\section{Compositional Image Retrieval}
\label{sup:compositional}
We further evaluate our model on the \emph{compositional image retrieval} task introduced in GT-Loc~\cite{Shatwell_2025_ICCV}, which tests whether a model can jointly reason about \emph{both} geographic location and time-of-capture.  
Given a query consisting of a location $l$ and a target time $t$, the goal is to retrieve an image that matches both attributes, \textit{i.e.}, an image captured \emph{at the same place} and \emph{at the target time}.  
This requires a unified representation where location and time interact meaningfully rather than being processed independently.
We follow the evaluation protocol in GT-Loc and benchmark on two datasets: \textsc{TIGeR-test-86k} and \textsc{CVT}.  
For each query, we create a multimodal query embedding corresponding to the location and time representations, and retrieve the nearest image embedding.  
A retrieved sample is correct if it matches the location of the query and query timestamp.

Across both datasets and all recall levels, \tiger delivers substantial improvements, achieving a \textbf{+3.9 to +14.7 absolute gain} over the SoTA baselines on R@1 (Table \ref{tab:composite-img-ret}).  
This shows that modeling cross-modal interactions is essential for accurate geo-temporal reasoning and enables significantly more reliable compositional retrieval.

\begin{table*}[h]
    \centering
    \setlength{\tabcolsep}{2pt}
    \renewcommand{\arraystretch}{1.0}
    \caption{\textbf{Compositional Image Retrieval on Unseen Cameras}. Given a query geo-location ($l$) and time ($t$), the task is to retrieve an image that matches the query’s geo-temporal attributes. Our model is trained with a contrastive loss, $\mathcal{L}_C(\bar{\mathbf{v}}, \bar{\mathbf{lt}})$, explicitly designed to align visual and fused geo-temporal modalities. This enables the model to learn rich interdependencies between images and their spatiotemporal context, allowing it to perform tasks that prior methods struggled to solve accurately. Across both the \textsc{TIGeR-test-86k} and CVT datasets, our approach achieves substantial improvements on all evaluation metrics. *Indicates methods we re-implemented following the protocols detailed in prior work.}

    \label{tab:composite-img-ret}
    \begin{tabular}{lccccccc} 
        \toprule
        \multirow{2}{*}{\textbf{Method}} & \multirow{2}{*}{\textbf{Retrieval}} & \multicolumn{3}{c}{\textbf{TIGeR-test-86k}} & \multicolumn{3}{c}{\textbf{CVT}} \\
        \cmidrule(lr){3-5} \cmidrule(lr){6-8}  
        &  &\textbf{R@1 (\%)} & \textbf{R@5 (\%)} & \textbf{R@10 (\%)} & \textbf{R@1 (\%)} & \textbf{R@5 (\%)} & \textbf{R@10 (\%)} \\
        
        \midrule
        
        Zhai et al.~\cite{zhai2019learning}* & $lt \rightarrow I$ & 6.63 & 19.17 & 26.99 & 2.14 & 6.30 & 9.81 \\
        Zhai et al. CLIP~\cite{zhai2019learning}* & $It \rightarrow I$ & 13.90 & 33.85 & 43.71 & 13.64 & 33.92 & 43.92 \\
        GT-Loc~\cite{Shatwell_2025_ICCV}* & $lt \rightarrow I$ & 3.18 & 12.11 & 19.08 & 23.33 & 49.07 & 60.86 \\
        \midrule
        \textbf{TIGeR (Ours)} & $lt \rightarrow I$ & \textbf{17.84} & \textbf{41.07} & \textbf{51.95} & \textbf{31.61} & \textbf{56.43} & \textbf{65.63} \\

        \bottomrule
    \end{tabular}
\end{table*}

\section{Analysis on geo-time-aware image retrieval}
\label{sup:gt-cond-img-ret}

Given a query image $I^Q$ and a target time $t^Q$, the goal is to retrieve a gallery image $I^G$ captured at (or near) the same geographic location as $I^Q$, but at the specified time $t^Q$. 
Solving this task requires the model to reason jointly about image, location, and time, rather than relying solely on visual semantics. 


We further analyze geo-time aware retrieval by comparing performance across the northern and southern hemispheres on \textsc{TIGeR-test-86k} and CVT.  
For every model and dataset, we compute Recall@10 separately for the northern (N-R@10) and southern (S-R@10) hemispheres, and report their ratio (N/S). 
Ratios closer to 1 indicate more balanced performance between hemispheres. 
As shown in Table~\ref{tab:geotime-hemispheres}, \tiger not only achieves the highest recall on both datasets, but also exhibits the most balanced N/S ratios, indicating the most geographically robust retrieval.

\begin{table}[h!]
\centering
\caption{\textbf{Hemispheric balance in geo-time-aware image retrieval.} 
We report Recall@10 on the northern (N-R@10) and southern (S-R@10) hemispheres, along with their ratio (N/S), for \textsc{TIGeR-test-86k} and CVT. 
Values of N/S closer to 1 indicate more balanced performance across hemispheres. 
\tiger attains both the highest recalls and the most balanced N/S ratios on both datasets.}
\label{tab:geotime-hemispheres}
\resizebox{\columnwidth}{!}{
\begin{tabular}{lcccccc}
\toprule
\multirow{2}{*}{\textbf{Method}} 
& \multicolumn{3}{c}{\textbf{TIGeR-Test-86k}} 
& \multicolumn{3}{c}{\textbf{CVT}} \\
\cmidrule(lr){2-4} \cmidrule(lr){5-7}
& \textbf{N-R@10} & \textbf{S-R@10} & \textbf{N/S}
& \textbf{N-R@10} & \textbf{S-R@10} & \textbf{N/S} \\
\midrule
Zhai et al.        & 8.47  & 14.12 & 0.60 & 3.35  & 1.46  & 2.29 \\
Zhai et al. CLIP   & 11.43 & 16.07 & 0.71 & 17.09 & 9.86  & 1.73 \\
GT-Loc             & 2.46  & 3.47  & 0.71 & 27.26 & 17.92 & 1.52 \\
\midrule
\textbf{TIGeR}     & \textbf{32.15} & \textbf{43.13} & \textbf{0.75} 
                   & \textbf{33.73} & \textbf{26.46} & \textbf{1.27} \\
\bottomrule
\end{tabular}
}
\end{table}

\section{Analysis on time prediction}
\label{sup:time-pred}

\begin{figure*}[h]
\begin{center}
\includegraphics[width=0.9\linewidth]{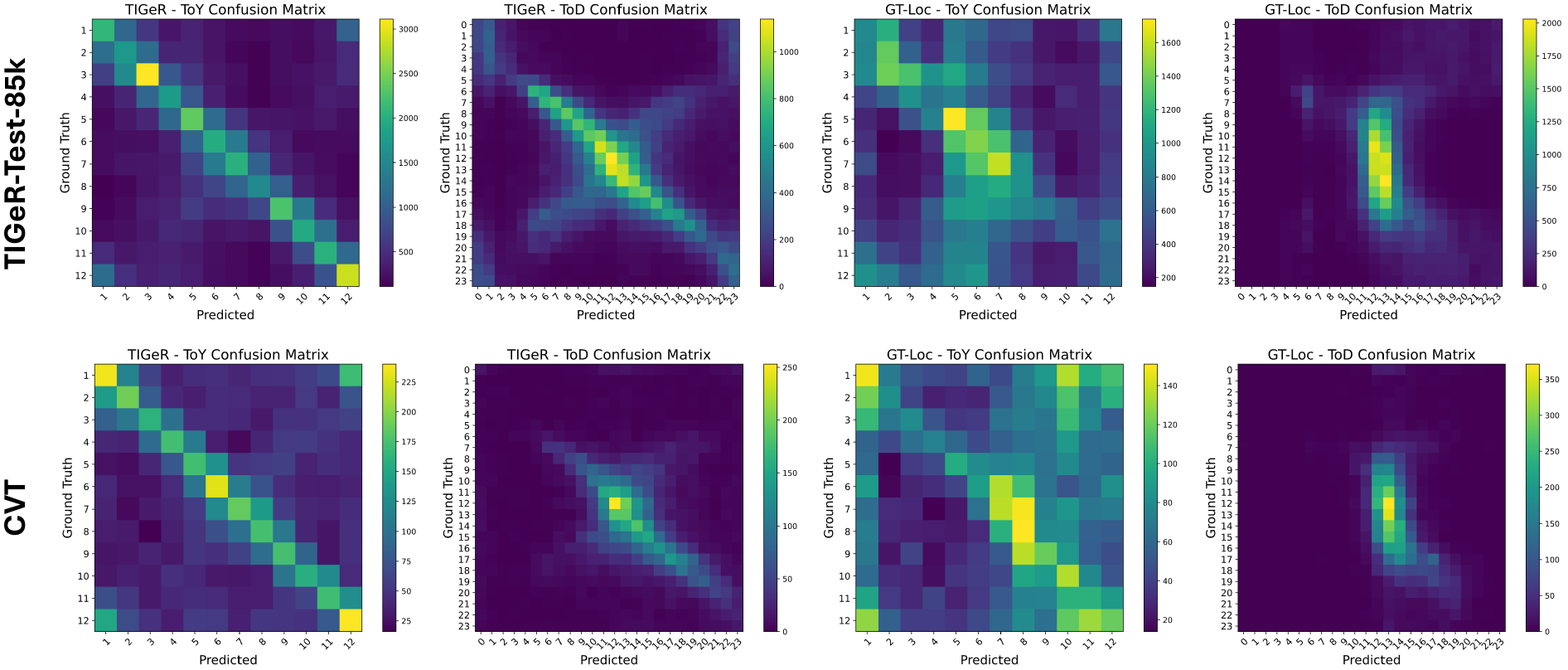}
\end{center}
\vspace{-1em}
\caption{\textbf{Confusion matrices} for \tiger and GT-Loc \cite{Shatwell_2025_ICCV} on the \textsc{TIGeR-test-86k} and CVT benchmarks. Ground-truth and predicted time-of-year (ToY) and time-of-day (ToD) are discretized into 12 monthly bins and 24 hourly bins, respectively. \tiger exhibits strong diagonal structure in both ToY and ToD matrices, indicating that predictions closely match the ground truth, with most errors occurring in adjacent temporal classes. This highlights \tiger’s substantially superior temporal-prediction performance compared to GT-Loc.
}
\label{fig:conf-mats}
\end{figure*}

In Figure \ref{fig:conf-mats}, we present the confusion matrices comparing \tiger and GT-Loc on both \textsc{TIGeR-test-86k} and CVT to evaluate the quality of our time-of-year (ToY) and time-of-day (ToD) predictions. A perfect temporal predictor produces a strictly diagonal confusion matrix, indicating that predicted time bins match the ground truth exactly.

Across both benchmarks, \tiger exhibits sharp, well-defined diagonals in both ToY and ToD matrices, with only small amounts of spread into adjacent bins. This pattern reflects precise temporal reasoning: predictions are correct or fall within the nearest neighboring month or hour, consistent with natural continuity in seasonal and lighting conditions.
In contrast, GT-Loc produces more off-diagonal mass, with predictions often drifting toward a narrow range of time bins, especially in the ToD setting. This indicates difficulty in capturing fine-grained temporal cues and a tendency to over-smooth predictions toward frequent modes.
Overall, the confusion matrices clearly show that \tiger delivers substantially stronger temporal predictions, maintaining tight alignment between predicted and true time bins and demonstrating markedly improved geo-temporal understanding compared to GT-Loc.

\section{Additional time-prediction baselines}

TICL~\cite{lin2025what} is a recent image–time pretraining approach that leverages contrastive learning to align image and temporal embeddings within a shared feature space, similar in spirit to GeoCLIP~\cite{vivanco2024geoclip} and GT-Loc~\cite{Shatwell_2025_ICCV}. The resulting pretrained model can be applied to downstream tasks such as time-based image retrieval, video scene classification, and time-aware image editing, demonstrating the benefits of temporally aware representations. For a fair comparison, we train TICL for joint time-of-year (ToY) and time-of-day (ToD) prediction on our proposed dataset, closely following the original training protocol. As shown in Table~\ref{tab:ticl-baseline}, TIGeR consistently achieves lower time prediction errors across all metrics on both the \textsc{TIGeR-test-86k} and CVT test sets, further supporting the conclusion that jointly modeling geo-location and time with early fusion yields more effective temporal representations.

\begin{table}[t]
\centering
\setlength{\aboverulesep}{0pt}
\setlength{\belowrulesep}{0pt}
\caption{\textbf{Time prediction with additional baselines}. We train and test TICL~\cite{lin2025what} on our proposed dataset, closely adhering to the training protocol outlined by the authors.}
\label{tab:ticl-baseline}
\begingroup
\setlength{\tabcolsep}{1pt}
\resizebox{\linewidth}{!}{
\begin{tabular}{lccccc} 
\toprule
\multirow{3}{*}{\textbf{Method}} & \multirow{3}{*}{\textbf{Retrieval}}
  & \multicolumn{2}{c}{\textbf{TIGeR-test-86k}}
  & \multicolumn{2}{c}{\textbf{CVT-test}} \\
\cmidrule(lr){3-4} \cmidrule(lr){5-6}
                & 
  & \textbf{ToY} & \textbf{ToD} 
  & \textbf{ToY} & \textbf{ToD}  \\
                & 
  & \textbf{Error}$\downarrow$ & \textbf{Error}$\downarrow$
  & \textbf{Error}$\downarrow$ & \textbf{Error}$\downarrow$  \\
\midrule

Zhai et al.~\cite{zhai2019learning}*                & $I \rightarrow t$ & 68.38 & 3.97 & 87.37 & 3.28  \\
Zhai et al. w/ CLIP ~\cite{zhai2019learning}*        & $I \rightarrow t$ & 57.51 & 3.22 & 68.95 & 2.8  \\
Time-Loc~\cite{Shatwell_2025_ICCV}*   & $I \rightarrow t$ & 74.87 & 4.02 & 65.10 & 2.86 \\
GT-Loc~\cite{Shatwell_2025_ICCV}*    & $I \rightarrow t$  & 74.58 & 3.52 & 78.95 & \textbf{2.68} \\
TICL~\cite{lin2025what}* & $I \rightarrow t$  & 59.88 & 3.47 & 72.63 & 3.27 \\

\midrule

\textbf{TIGeR (Ours)}    & $I \rightarrow t$  & \textbf{51.49} & \textbf{3.13} & \textbf{62.88} & 2.73 \\

\bottomrule
\end{tabular}}
\endgroup
\end{table}

\section{Training details}
\label{sup:training-details}

\paragraph{Optimization and schedule.}
We train TIGeR for 10k iterations with a global batch size of 1024. 
Validation on all benchmarks is performed every 500 iterations, corresponding to 20 training epochs over the effective number of iterations.
We use the AdamW optimizer with $(\beta_1,\beta_2) = (0.9, 0.999)$ and weight decay $10^{-3}$.
The learning rate follows a cosine decay schedule from a maximum value of $10^{-4}$ down to $10^{-7}$, with a linear warm-up phase during the first 100 iterations.
All experiments are run on a single NVIDIA H100 GPU with 80\,GB of memory, 88\,GB of system RAM, and a 12-core CPU.

\vspace{-1em}
\paragraph{Image encoder and augmentations.}
The image branch uses a pretrained ViT-L/14 CLIP encoder \cite{clip}, which is kept frozen during training and only serves to extract visual tokens as described in Section~4 of the main paper.
Input images are pre-processed with a standard CLIP-style pipeline.
Concretely, we apply a random resized crop to size $224\times224$ with scale sampled from $[0.6, 1.0]$ and aspect ratio from $[0.9, 1.1]$ using bicubic interpolation. 
We then convert the image to RGB, and with probability $0.25$ apply a mild color jitter (brightness, contrast, and saturation each sampled from $[-0.1, 0.1]$).
A random horizontal flip is applied with probability $0.5$.
Finally, images are converted to tensors and normalized with the mean and standard deviation used by CLIP.

\vspace{-1em}
\paragraph{Location and time encoders.}
Location and time are encoded with random Fourier feature (RFF) encoders, each followed by a small MLP.
GPS coordinates and temporal coordinates are first mapped from $\mathbb{R}^2$ into a 1024-dimensional feature space using a bank of sinusoidal projections with frequencies $\{2^0, 2^1, \dots, 2^9\}$.
The resulting RFF features are passed through a modality-specific MLP and layer normalization to produce the token sequences $L$ and $T$ used by the multimodal transformer.

\vspace{-1em}
\paragraph{Fusion transformer.}
The shared fusion module $\mathcal{F}(\cdot)$ is implemented as a single Transformer block that operates jointly on the concatenated tokens from the active modalities (image, location, and/or time).
The block uses 64 attention heads and the same hidden dimension as the encoder outputs (1024), and is shared across all six input configurations (single-modality and pair-wise modality inputs), as detailed in Section~4 of the main paper.
We found that a single well-parameterized block is sufficient to capture useful cross-modal interactions while keeping training efficient.

\begin{figure*}[t]
\begin{center}
\includegraphics[width=0.8\linewidth]{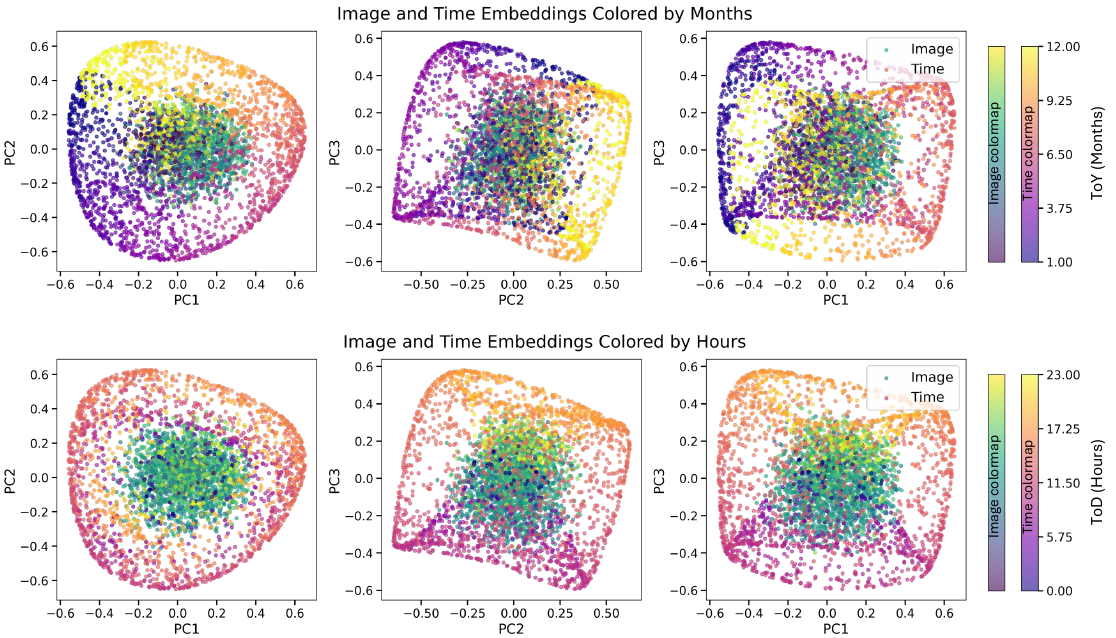}
\end{center}
\vspace{-1em}
\caption{\textbf{PCA visualization of the \tiger\ embedding space.}
We project 2{,}000 image--timestamp pairs from the CVT test set onto the first three principal components computed from the time embeddings, and apply the same projection to the corresponding image embeddings.
\textbf{Top:} embeddings colored by time-of-year (month). 
\textbf{Bottom:} embeddings colored by time-of-day (hour). 
Image embeddings (green--yellow) follow the toroidal structure traced by the time embeddings (purple--orange), showing that \tiger\ aligns visual and temporal features on a shared, torus-like manifold.
}
\label{fig:pca_supp}
\end{figure*}

\vspace{-1em}
\paragraph{Metric-aware classification heads.}
For geo-location classification, we discretize the Earth into HEALPix cells with $\text{NSIDE}=8$, yielding 768 equal-area regions.
For each training sample we compute the corresponding cell index and use the cell center as the class prototype.
Soft labels are constructed using the metric-affinity formulation from Eq.~(5) in the main paper, with Haversine distance as the metric and temperature $\gamma_{\text{geo}} = 250$.
For time, we discretize the flat torus $\mathbb{T}^2$ into $24\times 12 = 288$ bins corresponding to one-hour (ToD) and one-month (ToY) intervals.
Temporal soft labels are obtained using the torus geodesic distance with temperature $\gamma_{\text{time}} = 1$.
In both cases, a two-layer MLP applied to the pooled image embedding produces logits over the corresponding class space, which are trained with cross-entropy against the soft metric targets.

\vspace{-1em}
\paragraph{Training data and supervision.}
The main training signal comes from \textsc{TIGeR-Train-4.5M} and CVT, both of which provide image, geo-location, and timestamp annotations.
These datasets are used to optimize all components: contrastive losses over the unified embedding space, the multimodal fusion transformer, and the metric-aware classification heads.
In addition, we leverage auxiliary datasets that provide only images and location data (e.g., Google Landmarks v2 \cite{weyand2020google}, MP-16 \cite{7849098}, and OpenStreetViews 5M \cite{astruc2024openstreetview}).
For these datasets, we only train compute the loss using the image-location contrastive and classification losses using the same metric-aware targets.

\vspace{-1em}
\paragraph{Batch construction and debiasing.}
Some of the underlying datasets exhibit strong geographic imbalance, with a heavy skew toward North America and Europe.
To mitigate this, we enforce diversity at the batch level.
Each training batch of size 1024 is constructed by first sampling at least 64 distinct HEALPix cells at $\text{NSIDE}=8$.
For each selected cell, we then sample up to 16 images whose GPS coordinates fall inside that cell.
Whenever possible, we also enforce temporal diversity within each cell by avoiding repeated month--hour combinations; i.e., we try not to include two samples in the same batch that share the same (ToY, ToD) pair among the 288 possible configurations.
This strategy yields batches that are both geographically and temporally diverse, improving robustness and preventing the model from overfitting to overrepresented regions.

\vspace{-1em}
\paragraph{Entropy-adaptive reranking.}
At test time, we apply the entropy-adaptive reranking scheme described in Eqs.~(8)--(9) of the main paper to combine continuous retrieval scores with the classifier-based priors for geo-location and time.
For location retrieval, we use a similarity temperature $\psi_{\text{geo}} = 0.07$ and a maximum prior weight $\beta_{\text{geo}}^{\max} = 1$.
For time prediction, we set $\psi_{\text{time}} = 0.07$ and $\beta_{\text{time}}^{\max} = 2$.
The actual prior weight $\beta(I^Q)$ for a query image $I^Q$ is modulated by the entropy of the predicted class distribution, so that confident predictions rely more strongly on the prior, while uncertain ones fall back on pure similarity-based retrieval.

\vspace{-1em}
\paragraph{Baseline reproducibility.}
All baselines are reproduced as closely as possible by following the training details provided in their respective papers. For a fair comparison with our method, we train all baselines on the \textsc{TIGeR-test-86k} and CVT datasets. In the baseline variants where the original CNN backbone is replaced with a pretrained CLIP ViT, we also match the number of gradient updates used by our method to ensure a consistent and fair comparison.

\section{Embedding space visualization}
\label{sup:embedding}

To inspect the structure of the learned geo-temporal representation, we visualize a PCA projection of the joint embedding space (Figure \ref{fig:pca_supp}). We randomly sample 2{,}000 image-time pairs from the CVT test set and compute PCA on the time embeddings; the resulting projection is then applied to the corresponding image embeddings so that both modalities share the same principal-component basis. Image embeddings are shown with a green--yellow colormap, and time embeddings with a purple--orange colormap. We display the three pairwise combinations of the first three components (PC1--PC2, PC2--PC3, and PC1--PC3) to better reveal the global geometry of the space.

The top row colors each point by time-of-year (month). Both image and time embeddings trace a smooth circular pattern along the outer ring, with neighboring months occupying adjacent regions. The bottom row colors points by time-of-day (hour), revealing a complementary cycle that wraps around an inner ring. Taken together, these views indicate that \tiger organizes image and time features on a low-dimensional, torus-like manifold, consistent with our design choice of encoding time as a point on the flat torus~$\mathbb{T}^2$.